\documentclass[]{template/preprint}

\usepackage{booktabs}
\usepackage{multirow}
\usepackage{amsmath}
\usepackage{amssymb}
\usepackage[linesnumbered,ruled,vlined]{algorithm2e}
\usepackage{float}
\usepackage{graphicx}
\usepackage{adjustbox}
\usepackage{array}
\usepackage{tabularx}

\newcolumntype{Y}{>{\raggedright\arraybackslash}X}
\makeatletter
\newcommand{\appendixcontentsline}[5]{%
  \@dottedtocline{#1}{#2}{#3}{#4}{\pageref{#5}}%
}
\makeatother

\title{TorchCraft: Unified binder design by inverting an all-atom structure predictor}

\author[*]{TorchCraft Team}

\affiliation[]{Changping Laboratory, Beijing, China}

\abstract{
All-atom structure predictors model diverse molecular interactions, but using their learned
structural priors for binder design remains challenging. Here we present TorchCraft, a unified
binder-design framework that optimizes sequence logits through a frozen all-atom predictor.
Implemented in TorchFold, TorchCraft combines confidence, contact, geometric, and sequence-prior
objectives within a shared optimization procedure for minibinders, framework-conditioned VHHs,
cyclic peptides, and ligand-binding proteins. Using pretrained AlphaFold3 weights, TorchCraft generated representative minibinders and VHHs
with experimentally measured binding across four targets in each format, without post hoc sequence
redesign. Computational benchmarks further demonstrated the framework's applicability to cyclic
peptides and ligand-conditioned pocket design. TorchCraft
extends predictor inversion to multiple binder formats and molecular contexts, providing a common
framework for reusing all-atom structural priors in design.
}

\checkdata[Correspondence]{Mingchen Chen
(\url{mingchenchen@cpl.ac.cn})}
\checkdata[Project Page]{\url{https://torchx-cpl.github.io}}
\checkdata[Code]{\url{https://github.com/Mingchenchen/TorchX}}

\fancypagestyle{firststyle}{
    \fancyhead[L]{
        \vskip -2mm
    \includegraphics[height=16mm,keepaspectratio]{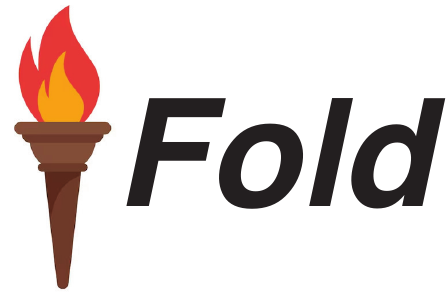}
    }
    \fancyhead[R]{}
    \fancyfoot[L]{%
      \begin{minipage}[b]{0.7\textwidth}
        \hrule height 0.4pt width 2in
        \vspace{4pt}
        \footnotesize $^*$Full author list in \hyperref[sec:author-contributions]{Author Contributions}.
      \end{minipage}%
    }
    \fancyfoot[C]{}
    \fancyfoot[R]{\thepage}
}

\begin{document}

\maketitle

\section{Introduction}

De novo binder design seeks sequences that fold reliably and recognize specified molecular
targets.\cite{watkinsStructurebasedInhibitionProtein2015,luRecentAdvancesDevelopment2020,khakzadNewAgeProtein2023}
Advances in structure prediction have made learned structural priors increasingly useful for this
task. AlphaFold2 models protein structure and assembly, whereas all-atom predictors such as
AlphaFold3 can additionally represent nucleic acids, small molecules, ions, and modified
residues.\cite{af2,af3,boltz2,protenix} These capabilities create an opportunity to guide binder
design within a common molecular representation, provided that the predictor's outputs can be
converted into effective sequence-optimization objectives.

Current design workflows combine structure generation, sequence assignment, and evaluation in
different ways. Backbone-generation methods, including RFdiffusion, are commonly paired with
ProteinMPNN or LigandMPNN and subsequent structure
prediction.\cite{rfdiffusion,rfdiffusion2,rfdiffusion3,proteinmpnn,ligandmpnn,BoltzGen} These
workflows provide complementary ways to explore sequence and structure, with task scope depending
on the molecular representations and constraints supported by each component.

Predictor-inversion methods instead optimize sequences directly against outputs of a pretrained
structure predictor.\cite{hallucination,af2inversion} BindCraft established an experimentally
effective AF2-based workflow for protein binder design.\cite{bindcraft} BoltzDesign extended
all-atom predictor inversion to additional molecular contexts, while Germinal incorporated
antibody-specific objectives and sequence guidance.\cite{boltzdesign,Germinal} Forward-pass
approaches such as Protein Hunter and HalluDesign offer a related strategy by alternating structure
hallucination and sequence redesign without backpropagating through the
predictor.\cite{proteinhunter,halludesign} These developments motivate further evaluation of how
all-atom predictors can support effective design across binder formats and target chemistries.

A practical question is whether a shared predictor-inversion framework can accommodate different
binder formats and molecular contexts while producing experimentally functional sequences.
Addressing this question requires both task-specific constraints, such as a fixed antibody
framework or peptide cyclization, and sequence-level controls that complement structural
objectives. It also requires evaluation beyond the confidence scores used during optimization.

Here we present TorchCraft, a binder-design framework that optimizes sequence logits through a
frozen all-atom structure predictor implemented in TorchFold.\cite{liu2026torchfold} All designs
reported here, including experimentally validated binders and computational benchmarks, were
generated using pretrained AF3 weights. The contribution is a shared optimization procedure with
task-specific sequence masks, restraints, priors, and schedules, together with experimental
minibinder and VHH demonstrations and computational extensions to cyclic peptides and
small-molecule-binding proteins.

\section{Results}

\subsection{TorchCraft as an all-atom hallucination framework}

\begin{figure}[H]
  \centering
  \IfFileExists{figs-TD/fig1.pdf}{
    \includegraphics[width=\textwidth]{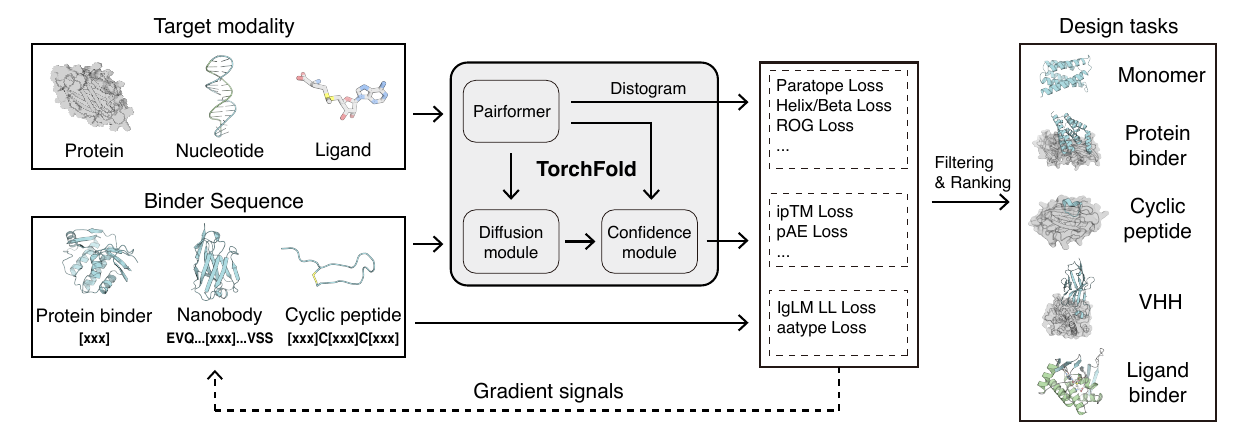}
  }{
    \fbox{\parbox[c][0.25\textheight][c]{0.92\textwidth}{
    \centering
    Missing figure file: figs-TD/fig1.pdf.\par
    TorchCraft overview: differentiable predictor-inversion workflow.
    }}
  }
  \caption{
    \textbf{TorchCraft optimizes binder sequences through a frozen all-atom
    predictor.} Target context and editable sequence representations are
    evaluated using the TorchFold implementation with pretrained AF3 weights,
    and gradients of task-specific objectives update the sequence logits.
    Experimental campaigns and computational benchmarks reported here used the
    same AF3-weight setting. The framework is applied to minibinders,
    framework-conditioned VHHs, cyclic peptides, and ligand-conditioned
    proteins by changing the editable positions, molecular constraints, and
    loss terms.
  }
  \label{fig:1}
\end{figure}

TorchCraft represents the editable binder sequence as a trainable logit matrix and updates it using
gradients of design objectives computed through a frozen predictor (Figure~\ref{fig:1}). Each
iteration evaluates the target--binder system and adjusts the sequence to improve predicted
structural compatibility, interface formation, and task-specific sequence properties. Optimization
progresses from continuous sequence representations to discrete amino-acid assignments, yielding a
designed sequence and a corresponding predicted all-atom structure (Figure~\ref{fig:2}b).

The all-atom representation allows the target context to include protein and nonprotein components.
In this study, we apply the framework to minibinders, framework-conditioned VHHs, head-to-tail and
disulfide-linked cyclic peptides, and proteins designed around small-molecule ligands. These tasks
share the sequence-optimization engine but use different editable positions, loss terms, molecular
constraints, and optimization schedules (Supplementary Table~\ref{tab:design_threeline}).

TorchCraft uses the TorchFold implementation to perform sequence optimization with a frozen
predictor. Computational optimizations, including conformer-generation parallelism, gradient
checkpointing, caching, and fused attention, support repeated evaluation during optimization
(Appendix~\ref{app:implementation-optimizations}).

\subsection{De novo design of protein binders against protein targets}

\begin{figure}[H]
  \centering
  \IfFileExists{figs-TD/fig2.pdf}{
    \includegraphics[width=0.9\textwidth]{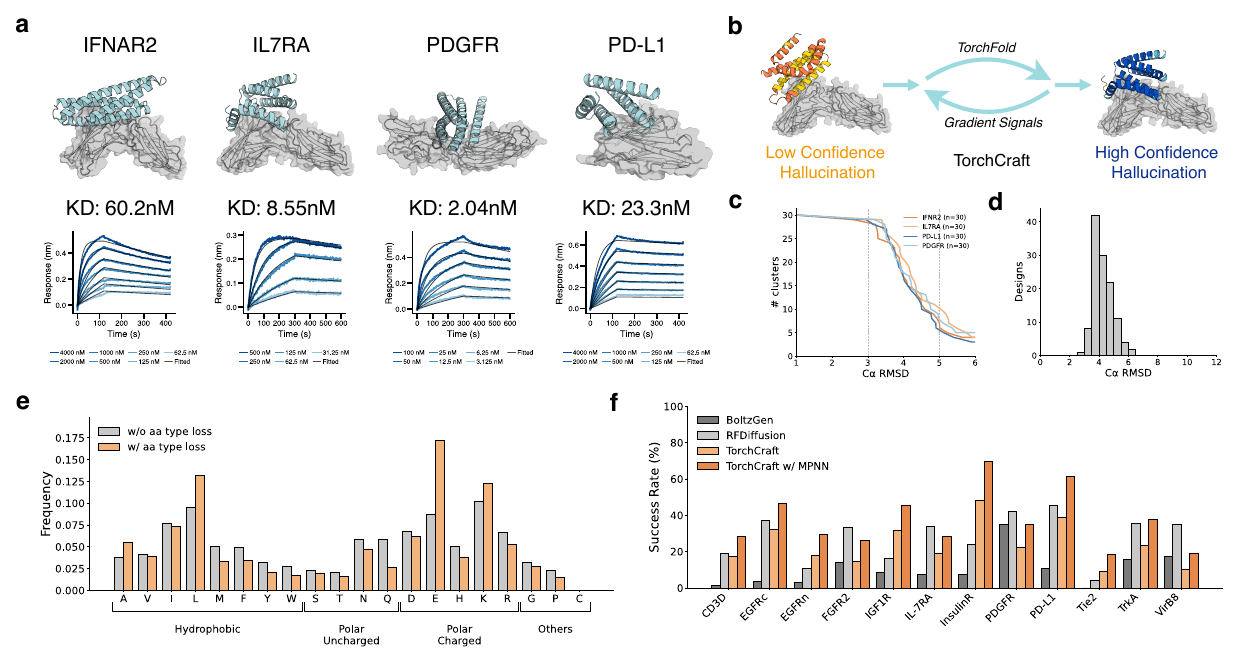}
  }{
    \fbox{\parbox[c][0.23\textheight][c]{0.92\textwidth}{
    \centering
    Missing figure file: figs-TD/fig2.pdf.\par
    Expected minibinder panels: experimental validation, design schematic,
    diversity, novelty, amino-acid composition, and benchmark.
    }}
  }
  \caption{
    \textbf{Experimental minibinder binding and computational evaluation of
    TorchCraft designs.} (a) Predicted complexes and BLI measurements for
    representative minibinders against IFNAR2, IL-7R$\alpha$, PDGFR$\beta$, and
    PD-L1. Apparent dissociation constants of the displayed binders were 60.2,
    8.55, 2.04, and 23.3~nM, respectively. (b) Schematic of sequence
    optimization, showing conversion of a continuous sequence representation into a
    discrete designed sequence and predicted complex through iterative
    predictor evaluation and sequence updates. (c,d) Diversity and divergence
    of predicted binding configurations for the analyzed designs, quantified
    using target-aligned binder C$\alpha$ RMSD clustering and comparison with
    interactions identified around similar target chains in the PDB.
    (e) Amino-acid composition of TorchCraft minibinders optimized with or
    without the composition regularizer, which matches the mean residue
    distribution to a ProteinMPNN-derived reference. (f) Computational pass
    rates across 12 protein targets, aggregated over binder lengths of 75, 120,
    160, and 200 residues. TorchCraft, BoltzGen, and RFdiffusion designs were
    evaluated with TorchScore, which scores supplied coordinates using the same
    pretrained AF3 weights used for design and does not refold sequences.
  }
  \label{fig:2}
\end{figure}

We designed minibinders against IFNAR2, IL-7R$\alpha$, PDGFR$\beta$, and PD-L1 using target
structures, binder-length ranges, and hotspot definitions specified in Supplementary
Table~\ref{tab:minibinder_validation_settings}. Candidates were selected from 3,000 generated
designs per target using the filters and ranking procedure in
Appendix~\ref{app:minibinder-candidate-selection}. Multi-concentration BLI measurements identified
binders in all four campaigns, with representative apparent dissociation constants of 60.2, 8.55,
2.04, and 23.3~nM, respectively (Figure~\ref{fig:2}a). The experimentally tested sequences were
obtained directly from TorchCraft without post hoc sequence redesign. Predicted complexes of the
analyzed designs displayed diverse binding configurations and differed from interactions identified
in the PDB comparison (Figure~\ref{fig:2}c,d).

Structural objectives do not directly constrain all sequence features relevant to expression. We
therefore added an amino-acid-composition regularizer that matches the average residue distribution
to a reference estimated from ProteinMPNN-redesigned samples. The regularizer altered the
composition of TorchCraft sequences (Figure~\ref{fig:2}e) and increased ESMFold confidence across
the 12-target benchmark (Supplementary Figure~\ref{fig:S1}a). In two PDGFR$\beta$ minibinder
campaigns, the round incorporating composition regularization showed higher expression than the
preceding unregularized round (Supplementary Figure~\ref{fig:S1}b).

We next compared TorchCraft with BoltzGen and RFdiffusion across 12 protein targets and binder
lengths of 75, 120, 160, and 200 residues. Designs were evaluated using TorchScore, a score-only
AF3 evaluation of supplied coordinates
(Appendix~\ref{app:torchscore}). Computational pass rates varied by
target and binder length (Figure~\ref{fig:2}f). Raw TorchCraft designs were competitive on several targets,
while optional ProteinMPNN redesign of noninterface residues further improved pass rates in many
settings. Because the evaluation uses predictor-derived scores, we additionally report Rosetta
interface-score distributions (Supplementary Figure~\ref{fig:S2}).

\subsection{De novo VHH design by framework-conditioned CDR optimization}

\begin{figure}[H]
  \centering
  \IfFileExists{figs-TD/fig3.pdf}{
    \includegraphics[width=0.9\textwidth]{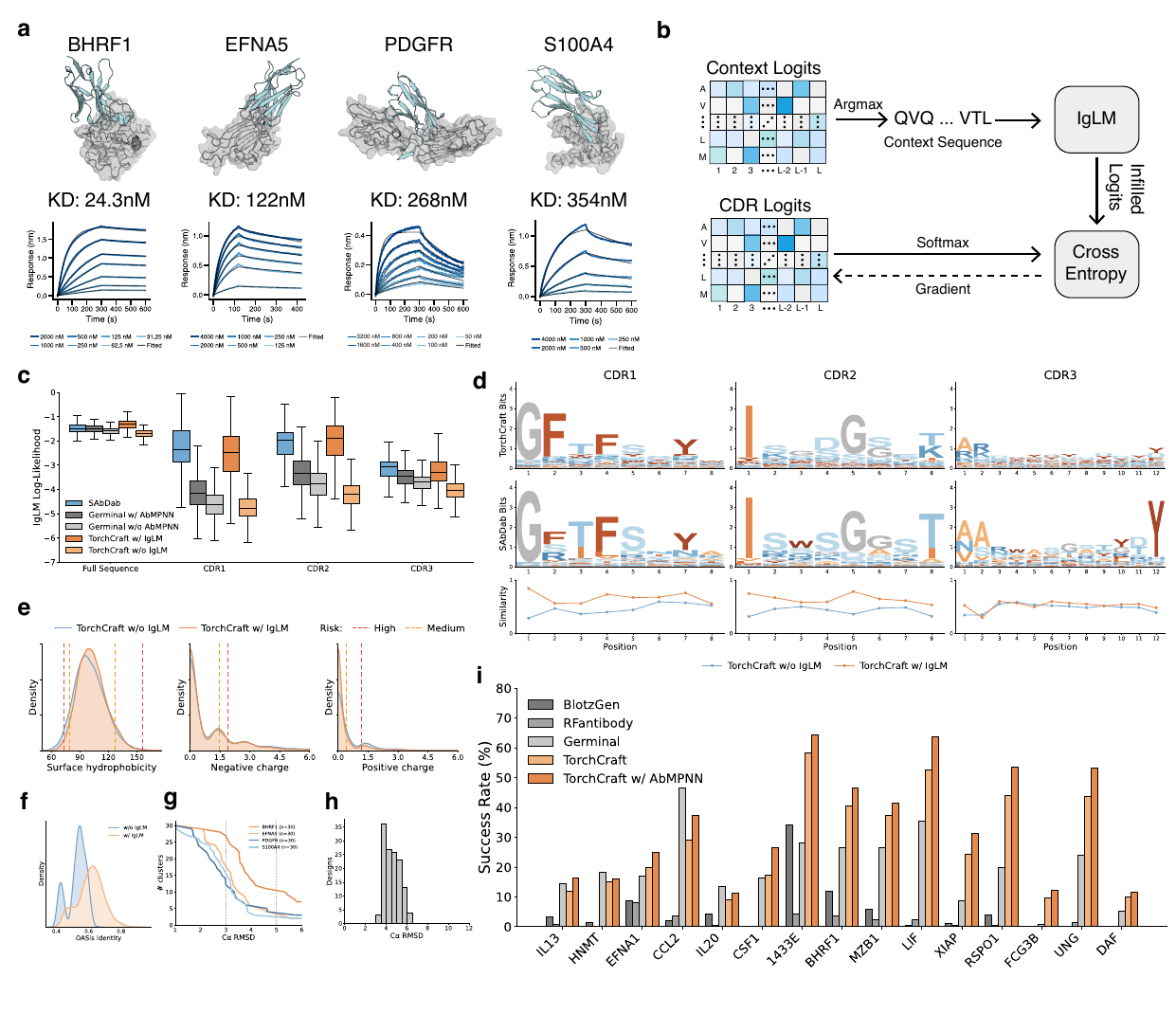}
  }{
    \fbox{\parbox[c][0.23\textheight][c]{0.92\textwidth}{
    \centering
    Missing figure file: figs-TD/fig3.pdf.\par
    Expected VHH panels: experimental validation, IgLM guidance, sequence
    analysis, developability, humanness, diversity, novelty, and benchmark.
    }}
  }
  \caption{
    \textbf{Framework-conditioned VHH design with CDR-specific sequence
    guidance.} (a) Predicted complexes and BLI measurements for representative
    VHHs against BHRF1, EFNA5, PDGFR$\beta$, and S100A4. Apparent dissociation
    constants of the displayed binders were 24.3, 122, 268, and 354~nM,
    respectively.
    (b) Schematic of CDR-restricted IgLM guidance: for each CDR, the framework
    and the remaining CDRs provide infilling context, and the IgLM-derived
    objective is applied only to the editable loop. (c) IgLM log-likelihood
    distributions for natural SAbDab VHHs, Germinal-generated sequences with or
    without CDR redesign, and TorchCraft sequences with or without
    CDR-restricted IgLM regularization. (d) Information-content sequence logos
    for HCDR1, HCDR2, and HCDR3, shown together with position-wise similarity to
    the SAbDab reference set. (e) Therapeutic Nanobody Profiler (TNP)
    comparison of developability-associated profiles for sequences generated
    with or without IgLM guidance.\cite{TNP} (f) BioPhi/OASis humanness-score
    comparison of guided and unguided sequences.\cite{BioPhi} (g,h) Diversity
    and divergence of predicted VHH
    binding configurations, quantified using target-aligned binder C$\alpha$
    RMSD clustering and comparison with PDB interactions. (i) Computational
    pass rates for VHH design across 15 targets, comparing TorchCraft with
    RFantibody, BoltzGen, and Germinal under a joint TorchScore-confidence and
    ESM2-perplexity criterion. TorchScore scores supplied coordinates using the
    same pretrained AF3 weights used for design and does not refold sequences.
    A framework-stratified analysis is shown in
    Supplementary Figure~\ref{fig:S5}.
  }
  \label{fig:3}
\end{figure}

We next applied TorchCraft to VHH design, holding a predefined framework sequence fixed while
optimizing the three CDRs. This confines sequence search to the paratope-forming regions while
retaining the selected scaffold context. Designs were generated against BHRF1, EFNA5, PDGFR$\beta$,
and S100A4 using the task specifications in Supplementary
Table~\ref{tab:vhh_validation_settings}. BLI measurements identified binders in all four campaigns,
with representative apparent dissociation constants of 24.3, 122, 268, and 354~nM, respectively
(Figure~\ref{fig:3}a). These sequences were tested without post hoc sequence redesign. The
corresponding predicted complexes were analyzed for diversity and divergence from PDB binding
configurations (Figure~\ref{fig:3}g,h).

Unregularized VHH optimization produced CDR sequences with residue preferences that differed from
the SAbDab reference set. To incorporate antibody-sequence constraints during design, we added an
IgLM-derived objective evaluated separately for CDR1, CDR2, and CDR3.\cite{shuaiIgLMInfillingLanguage2023} For each CDR, the framework
and the current sequences of the other CDRs provide context for the local infilling objective
(Figure~\ref{fig:3}b; Algorithm~\ref{alg:iglm_cdr}). This concentrates sequence guidance on the
editable regions. We compared this approach with the full-sequence guidance used in the Germinal
implementation evaluated here.\cite{Germinal}

CDR-restricted IgLM guidance increased IgLM likelihoods and shifted position-specific residue
preferences toward those observed in the SAbDab reference set, particularly in CDR1 and CDR2
(Figure~\ref{fig:3}c,d). Guided sequences also showed changes in TNP developability-associated
profiles and higher OASis humanness scores (Figure~\ref{fig:3}e,f).\cite{TNP,BioPhi} These analyses
indicate improved agreement with selected sequence-based reference distributions.

We compared TorchCraft with RFantibody, BoltzGen, and Germinal across 15 targets and five VHH
frameworks (Figure~\ref{fig:3}i; Supplementary
Figure~\ref{fig:S5}).\cite{rfantibody,BoltzGen,Germinal} Designs were assessed using the joint
TorchScore-confidence and ESM2-perplexity criterion described in
Appendix~\ref{app:benchmark-settings}. TorchCraft achieved higher computational pass rates on many
targets, with additional gains from optional AbMPNN redesign in several settings. Rosetta
interface-score distributions provide an additional assessment of the predicted complexes
(Supplementary Figure~\ref{fig:S3}).

\subsection{In silico benchmark and analysis of cyclic peptide design}

\begin{figure}[H]
  \centering
  \IfFileExists{figs-TD/fig4.pdf}{
    \includegraphics[width=0.9\textwidth]{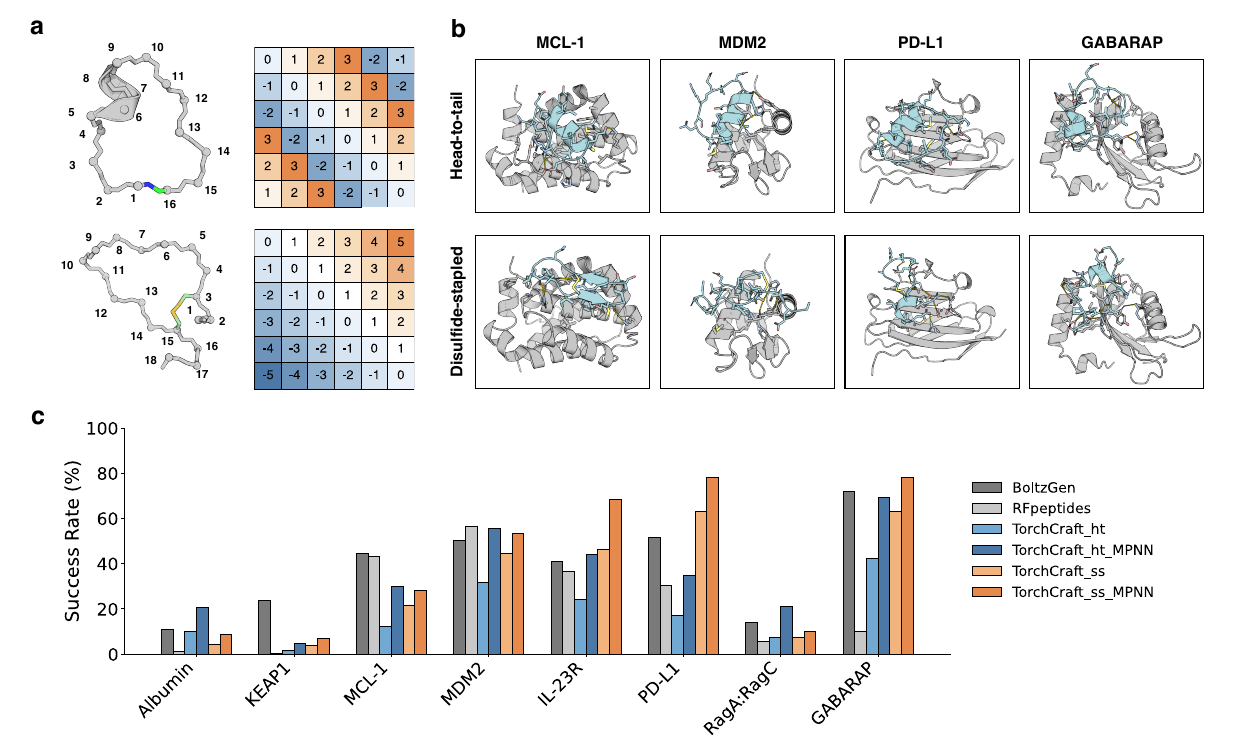}
  }{
    \fbox{\parbox[c][0.25\textheight][c]{0.92\textwidth}{
    \centering
    Missing figure file: figs-TD/fig4.pdf.\par
    Cyclic peptide design: cyclization schemes, representative designs, and
    benchmark performance.
    }}
  }
  \caption{
    \textbf{Computational evaluation of head-to-tail and disulfide-linked
    cyclic-peptide designs.} (a) Schematic comparison of the two
    parameterizations: head-to-tail designs use cyclic relative positional
    encoding to represent sequence adjacency across the termini, whereas
    disulfide-linked designs retain linear positional encoding and fix the
    cysteine positions that define the intended linkage. (b) Representative
    predicted complexes for MCL-1, MDM2, PD-L1, and GABARAP, shown for both
    cyclization modes. (c) Computational pass rates across eight target
    systems: MCL-1, MDM2, PD-L1, IL-23R, albumin, KEAP1, GABARAP, and
    RagA:RagC. BoltzGen, RFpeptides, and TorchCraft designs were compared using
    the structural and confidence criteria defined in
    Appendix~\ref{app:benchmark-settings}.
  }
  \label{fig:4}
\end{figure}

We next evaluated cyclic-peptide design computationally using two parameterizations
(Figure~\ref{fig:4}a). Head-to-tail designs used cyclic relative positional encoding to represent
sequence adjacency across the termini. Disulfide-linked designs retained linear positional encoding
and fixed the cysteine positions defining the intended linkage. Predicted structures were then
classified according to whether they satisfied the intended cyclization and backbone-integrity
criteria (Supplementary Figure~\ref{fig:S6}). Both settings used contact-based objectives together
with the additional terms specified in Supplementary Table~\ref{tab:design_threeline}.

We compared cyclic-peptide designs for eight target systems with outputs from RFpeptides and
BoltzGen using the structural and confidence criteria defined in
Appendix~\ref{app:benchmark-settings} (Figure~\ref{fig:4}c). Performance depended on both the
target and cyclization mode. Disulfide-linked TorchCraft designs showed higher computational pass
rates in several settings, including PD-L1 and GABARAP, whereas head-to-tail designs displayed a
different target-dependent profile. Optional ProteinMPNN redesign improved pass rates in several
comparisons. Representative predicted complexes and Rosetta interface-score distributions
illustrate the resulting interfaces (Figure~\ref{fig:4}b; Supplementary Figure~\ref{fig:S7}).

\subsection{Computational design of ligand-binding protein pockets}

\begin{figure}[H]
  \centering
  \IfFileExists{figs-TD/fig5.pdf}{
    \includegraphics[width=0.9\textwidth]{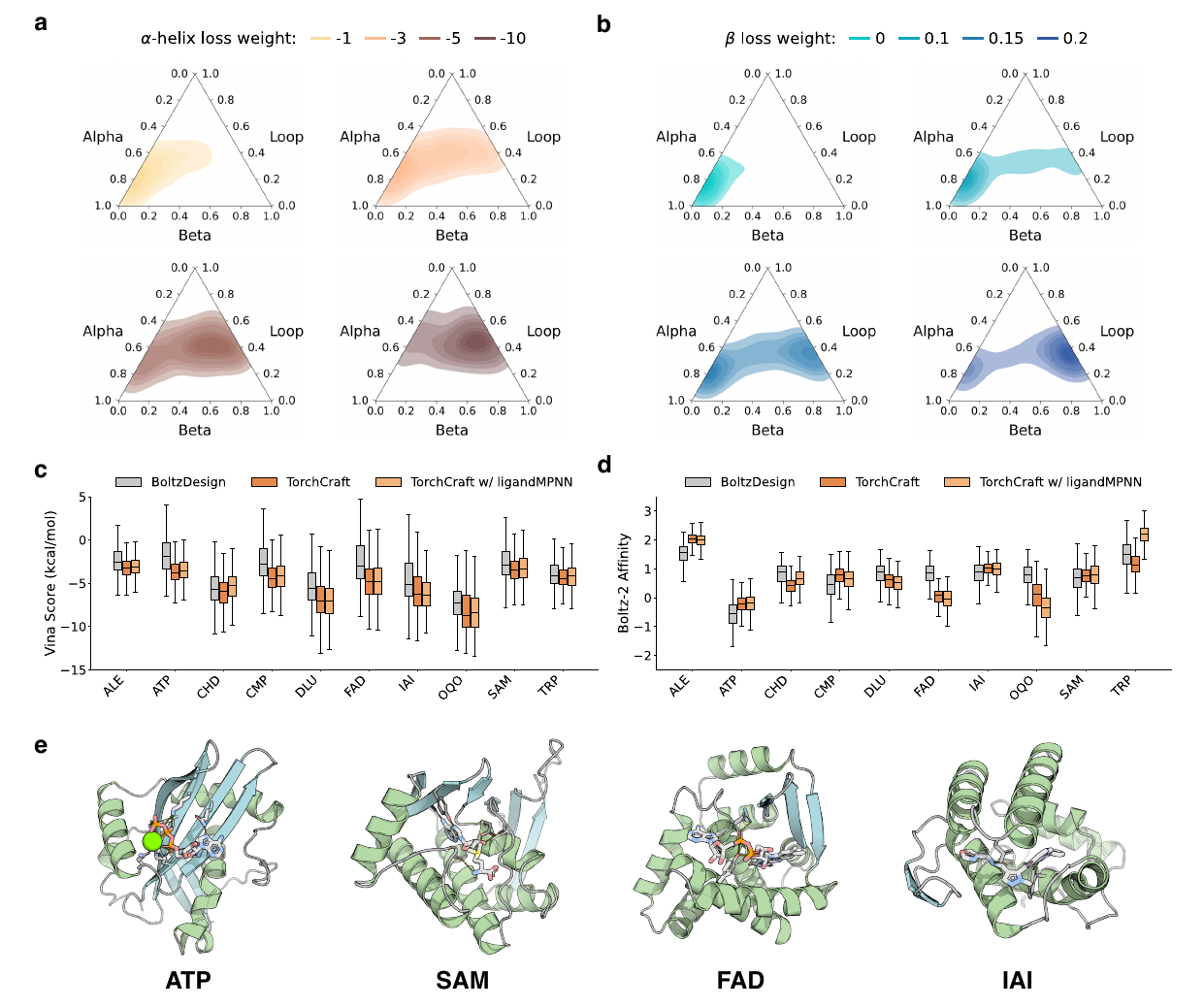}
  }{
    \fbox{\parbox[c][0.25\textheight][c]{0.92\textwidth}{
    \centering
    Missing figure file: figs-TD/fig5.pdf.\par
    Small-molecule interaction story: secondary-structure loss tuning, Vina and
    Boltz-2 benchmarks, and representative 10-ligand TorchCraft binders.
    }}
  }
  \caption{
    \textbf{Computational generation and evaluation of ligand-conditioned
    protein pockets.} (a,b) Helix and $\beta$-structure terms modulate the
    secondary-structure composition of designed binders. Ternary kernel density
    plots show compositions under different loss-weighting schemes. (c)
    AutoDock Vina docking scores (kcal~mol$^{-1}$) for designed pockets across
    the 10-ligand panel. (d) Boltz-2 affinity predictions for the same designs.
    Score distributions vary by ligand and method. (e) Representative predicted
    160-residue binders for four ligands: ATP, SAM, FAD, and IAI. Designs were
    generated without a supplied native scaffold. Raw outputs were evaluated
    without post hoc sequence redesign. An optional LigandMPNN-redesigned
    setting is analyzed separately.
  }
  \label{fig:5}
\end{figure}

Designing proteins around small-molecule ligands requires compatible folding, pocket geometry, and
local chemical interactions. We therefore combined ligand-contact objectives with controls on
binder compactness and secondary-structure composition. An isotropy regularizer was used to bias
the overall binder shape, while helix and $\beta$-structure terms modulated secondary-structure
preferences (Figure~\ref{fig:5}a,b). These objectives were intended to support pocket formation
without specifying a pre-existing scaffold.

Small-molecule design began with a ligand-free warm-up stage that optimized the binder before
ligand-dependent objectives were introduced. The ligand was then added, and the weight of the
interface objective was gradually increased before the final optimization stages
(Appendix~\ref{app:target-specific-stage-configurations}). This schedule separates initial fold
formation from subsequent pocket optimization.

We evaluated a panel of ten small-molecule targets spanning different chemical structures,
including nucleotides, cofactors, metabolites, and drug-like compounds (Supplementary
Figure~\ref{fig:S8}).\cite{boltzdesign,complexa,rosettafoldaa,halludesign} TorchCraft generated 500
independent 160-residue proteins per ligand without a supplied native scaffold. Raw outputs were
evaluated without post hoc sequence redesign or score-based preselection. An optional
LigandMPNN-redesigned setting was analyzed separately. Representative predicted pockets are shown
in Figure~\ref{fig:5}e.

We evaluated the predicted ligand-binding proteins using AutoDock Vina scores and Boltz-2 affinity
predictions (Figure~\ref{fig:5}c,d),\cite{AutoDock2010,AutoDock2021,boltz2} with TorchScore pass
rates reported separately (Supplementary Figure~\ref{fig:S9}). Score distributions varied across
ligands and methods. These complementary computational assessments support the feasibility of
generating ligand-conditioned pockets.

\section{Discussion}

TorchCraft extends the predictor-inversion approach to binder design within an all-atom molecular
representation. Building on the principle established by BindCraft and related
methods,\cite{bindcraft,boltzdesign} it uses a common optimization engine with task-specific
constraints for minibinders, VHHs, cyclic peptides, and ligand-conditioned proteins. Experiments
using pretrained AF3 weights established binding by designed minibinders and VHHs without post hoc
sequence redesign. These results show that direct sequence optimization can produce experimentally
functional binders in distinct protein formats, while computational studies support extension to
additional molecular contexts.

A practical advantage of TorchCraft is that design tasks can be specified through editable
residues, molecular context, restraints, and objective functions within the same framework. This
allows new constraints to be evaluated without training a separate generator for every task. The
present results support this approach across several design settings, although each requires
appropriate objectives, sequence priors, and validation. Optional sequence redesign remains useful
in some applications, even though it was not required for the experimentally tested binders
reported here.

The current study has several limitations. Experimental validation is confined to minibinders and
VHHs, whereas cyclic-peptide and small-molecule results remain computational. Predictor-derived
pass rates are sensitive to the evaluator and should not be interpreted as experimental success
rates. The functional consequences of sequence regularization also require matched experimental
comparisons, and binding measurements do not by themselves establish the accuracy of the predicted
interfaces. Further validation across targets, formats, and independent assays will clarify the
framework's practical range.

A further methodological direction is to incorporate objectives defined directly on atomic
interactions. Stable differentiation through coordinate generation could enable more explicit
control of ligand-contact geometry, buried polar networks, and coordination sites. Such objectives
would complement the confidence- and distogram-based signals used here, but their value must be
assessed through both geometric validation and experimental function.

TorchFold's reported improvements in antibody--antigen structure prediction provide a promising
direction for antibody design.\cite{liu2026torchfold} Its specialized checkpoint may offer more
informative guidance during CDR optimization. Matched
comparisons of experimental binding and expression under the same targets, frameworks, and
candidate-selection procedures will be needed to test this potential.

\clearpage

\section{Acknowledgments}

This work was funded by Changping Laboratory (CPNL project 2026D-03-01).

\section{Author Contributions}
\label{sec:author-contributions}

Yu Liu$^{1,*}$, Zhouhanyu Shen$^{1,*}$, Zhengyi Li$^{1}$, Xikun Huang$^{1}$, Jiaqi Liu$^{1}$, Shuxian Gao$^{2}$, Qilin Yu$^{1}$, Xiayan Qin$^{1}$, Yucheng Zhang$^{2}$, Mingchen Chen$^{1,\dag}$

\vspace{0.5em}
\noindent $^1$Changping Laboratory, Beijing, China

\noindent $^2$Huawei Technologies Co., Ltd.

\vspace{0.5em}
\noindent $^*$Equal contribution

\noindent $^\dag$Corresponding author

\vspace{0.5em}
\noindent Y.L. and Z.S. contributed equally to this work. M.C. designed research; Y.L., Z.S., and X.H. performed research and analyzed data; Z.L. and S.G. performed code engineering; J.L. performed computational cyclic-peptide analysis; Q.Y., X.Q., and Y.Z. participated in discussions; and Y.L. wrote the paper.

\section{Data Availability}

The TorchCraft inference implementation and design protocols are available at \url{https://github.com/Mingchenchen/TorchX}. The project page is available at \url{https://torchx-cpl.github.io}.

\clearpage

\bibliographystyle{unsrt}
\bibliography{TorchDesign}

\begin{thebibliography}{10}

\bibitem{watkinsStructurebasedInhibitionProtein2015}
A.~M. Watkins and P.~S. Arora.
\newblock Structure-based inhibition of protein–protein interactions.
\newblock {\em European Journal of Medicinal Chemistry}, 94:480--488, 2015.

\bibitem{luRecentAdvancesDevelopment2020}
H.~Lu, Q.~Zhou, J.~He, Z.~Jiang, C.~Peng, R.~Tong, and J.~Shi.
\newblock Recent advances in the development of protein–protein interactions
  modulators: {{Mechanisms}} and clinical trials.
\newblock {\em Sig Transduct Target Ther}, 5(1):213, 2020.

\bibitem{khakzadNewAgeProtein2023}
H.~Khakzad, I.~Igashov, A.~Schneuing, C.~Goverde, M.~Bronstein, and B.~Correia.
\newblock A new age in protein design empowered by deep learning.
\newblock {\em Cell Systems}, 14(11):925--939, 2023.

\bibitem{af2}
J.~Jumper, R.~Evans, A.~Pritzel, T.~Green, M.~Figurnov, O.~Ronneberger,
  K.~Tunyasuvunakool, R.~Bates, A.~Žídek, A.~Potapenko, A.~Bridgland,
  C.~Meyer, S.~A.~A. Kohl, A.~J. Ballard, A.~Cowie, B.~Romera-Paredes,
  S.~Nikolov, R.~Jain, J.~Adler, T.~Back, S.~Petersen, D.~Reiman, E.~Clancy,
  M.~Zielinski, M.~Steinegger, M.~Pacholska, T.~Berghammer, S.~Bodenstein,
  D.~Silver, O.~Vinyals, A.~W. Senior, K.~Kavukcuoglu, P.~Kohli, and
  D.~Hassabis.
\newblock Highly accurate protein structure prediction with {{AlphaFold}}.
\newblock {\em Nature}, 596(7873):583--589, 2021.

\bibitem{af3}
J.~Abramson, J.~Adler, J.~Dunger, R.~Evans, T.~Green, A.~Pritzel,
  O.~Ronneberger, L.~Willmore, A.~J. Ballard, J.~Bambrick, S.~W. Bodenstein,
  D.~A. Evans, C.-C. Hung, M.~O’Neill, D.~Reiman, K.~Tunyasuvunakool, Z.~Wu,
  A.~Žemgulytė, E.~Arvaniti, C.~Beattie, O.~Bertolli, A.~Bridgland,
  A.~Cherepanov, M.~Congreve, A.~I. Cowen-Rivers, A.~Cowie, M.~Figurnov, F.~B.
  Fuchs, H.~Gladman, R.~Jain, Y.~A. Khan, C.~M.~R. Low, K.~Perlin,
  A.~Potapenko, P.~Savy, S.~Singh, A.~Stecula, A.~Thillaisundaram, C.~Tong,
  S.~Yakneen, E.~D. Zhong, M.~Zielinski, A.~Žídek, V.~Bapst, P.~Kohli,
  M.~Jaderberg, D.~Hassabis, and J.~M. Jumper.
\newblock Accurate structure prediction of biomolecular interactions with
  {{AlphaFold}} 3.
\newblock {\em Nature}, 630(8016):493--500, 2024.

\bibitem{boltz2}
S.~Passaro, G.~Corso, J.~Wohlwend, M.~Reveiz, S.~Thaler, V.~R. Somnath,
  N.~Getz, T.~Portnoi, J.~Roy, H.~Stark, D.~Kwabi-Addo, D.~Beaini, T.~Jaakkola,
  and R.~Barzilay.
\newblock Boltz-2: {{Towards Accurate}} and {{Efficient Binding Affinity
  Prediction}}.
\newblock {\em bioRxiv}, 2025.

\bibitem{protenix}
{ByteDance AML AI4Science Team}, X.~Chen, Y.~Zhang, C.~Lu, W.~Ma, J.~Guan,
  C.~Gong, J.~Yang, H.~Zhang, K.~Zhang, S.~Wu, K.~Zhou, Y.~Yang, Z.~Liu,
  L.~Wang, B.~Shi, S.~Shi, and W.~Xiao.
\newblock Protenix - {{Advancing Structure Prediction Through}} a
  {{Comprehensive AlphaFold3 Reproduction}}, 2025.

\bibitem{rfdiffusion}
J.~L. Watson, D.~Juergens, N.~R. Bennett, B.~L. Trippe, J.~Yim, H.~E. Eisenach,
  W.~Ahern, A.~J. Borst, R.~J. Ragotte, L.~F. Milles, B.~I.~M. Wicky,
  N.~Hanikel, S.~J. Pellock, A.~Courbet, W.~Sheffler, J.~Wang, P.~Venkatesh,
  I.~Sappington, S.~V. Torres, A.~Lauko, V.~De~Bortoli, E.~Mathieu,
  S.~Ovchinnikov, R.~Barzilay, T.~S. Jaakkola, F.~DiMaio, M.~Baek, and
  D.~Baker.
\newblock De novo design of protein structure and function with
  {{RFdiffusion}}.
\newblock {\em Nature}, 620(7976):1089--1100, 2023.

\bibitem{rfdiffusion2}
W.~Ahern, J.~Yim, D.~Tischer, S.~Salike, S.~M. Woodbury, D.~Kim, I.~Kalvet,
  Y.~Kipnis, B.~Coventry, H.~R. Altae-Tran, M.~S. Bauer, R.~Barzilay, T.~S.
  Jaakkola, R.~Krishna, and D.~Baker.
\newblock Atom-level enzyme active site scaffolding using {{RFdiffusion2}}.
\newblock {\em Nat Methods}, 23(1):96--105, 2026.

\bibitem{rfdiffusion3}
J.~Butcher, R.~Krishna, R.~Mitra, R.~I. Brent, Y.~Li, N.~Corley, P.~T. Kim,
  J.~Funk, S.~Mathis, S.~Salike, A.~Muraishi, H.~Eisenach, E.~Sehgal,
  B.~Coventry, O.~Zhang, B.~Qiang, K.~Didi, F.~DiMaio, and D.~Baker.
\newblock De novo {{Design}} of {{All-atom Biomolecular Interactions}} with
  {{RFdiffusion3}}.

\bibitem{proteinmpnn}
J.~Dauparas, I.~Anishchenko, N.~Bennett, H.~Bai, R.~J. Ragotte, L.~F. Milles,
  B.~I.~M. Wicky, A.~Courbet, R.~J. De~Haas, N.~Bethel, P.~J.~Y. Leung, T.~F.
  Huddy, S.~Pellock, D.~Tischer, F.~Chan, B.~Koepnick, H.~Nguyen, A.~Kang,
  B.~Sankaran, A.~K. Bera, N.~P. King, and D.~Baker.
\newblock Robust deep learning–based protein sequence design using
  {{ProteinMPNN}}.
\newblock {\em Science}, 378(6615):49--56, 2022.

\bibitem{ligandmpnn}
J.~Dauparas, G.~R. Lee, R.~Pecoraro, L.~An, I.~Anishchenko, C.~Glasscock, and
  D.~Baker.
\newblock Atomic context-conditioned protein sequence design using
  {{LigandMPNN}}.
\newblock {\em Nat Methods}, 22(4):717--723, 2025.

\bibitem{BoltzGen}
H.~Stark, F.~Faltings, M.~Choi, Y.~Xie, E.~Hur, T.~O’Donnell, A.~Bushuiev,
  T.~Uçar, S.~Passaro, W.~Mao, M.~Reveiz, R.~Bushuiev, T.~Pluskal, J.~Sivic,
  K.~Kreis, A.~Vahdat, S.~Ray, J.~T. Goldstein, A.~Savinov, J.~A. Hambalek,
  A.~Gupta, D.~A. Taquiri-Diaz, Y.~Zhang, A.~K. Hatstat, A.~Arada, N.~H. Kim,
  E.~Tackie-Yarboi, D.~Boselli, L.~Schnaider, C.~C. Liu, G.-W. Li, D.~Hnisz,
  D.~M. Sabatini, W.~F. DeGrado, J.~Wohlwend, G.~Corso, R.~Barzilay, and
  T.~Jaakkola.
\newblock Toward {{Universal Binder Design}}.

\bibitem{hallucination}
I.~Anishchenko, S.~J. Pellock, T.~M. Chidyausiku, T.~A. Ramelot,
  S.~Ovchinnikov, J.~Hao, K.~Bafna, C.~Norn, A.~Kang, A.~K. Bera, F.~DiMaio,
  L.~Carter, C.~M. Chow, G.~T. Montelione, and D.~Baker.
\newblock De novo protein design by deep network hallucination.
\newblock {\em Nature}, 600(7889):547--552, 2021.

\bibitem{af2inversion}
C.~A. Goverde, B.~Wolf, H.~Khakzad, S.~Rosset, and B.~E. Correia.
\newblock De novo protein design by inversion of the {{{\textsc{AlphaFold}}}}
  structure prediction network.
\newblock {\em Protein Science}, 32(6):e4653, 2023.

\bibitem{bindcraft}
M.~Pacesa, L.~Nickel, C.~Schellhaas, J.~Schmidt, E.~Pyatova, L.~Kissling,
  P.~Barendse, J.~Choudhury, S.~Kapoor, A.~Alcaraz-Serna, Y.~Cho, K.~H.
  Ghamary, L.~Vinué, B.~J. Yachnin, A.~M. Wollacott, S.~Buckley, A.~H.
  Westphal, S.~Lindhoud, S.~Georgeon, C.~A. Goverde, G.~N. Hatzopoulos,
  P.~Gönczy, Y.~D. Muller, G.~Schwank, D.~C. Swarts, A.~J. Vecchio, B.~L.
  Schneider, S.~Ovchinnikov, and B.~E. Correia.
\newblock One-shot design of functional protein binders with {{BindCraft}}.
\newblock {\em Nature}, 646(8084):483--492, 2025.

\bibitem{boltzdesign}
Y.~Cho, M.~Pacesa, Z.~Zhang, B.~E. Correia, and S.~Ovchinnikov.
\newblock Boltzdesign1: {{Inverting All-Atom Structure Prediction Model}} for
  {{Generalized Biomolecular Binder Design}}.
\newblock {\em bioRxiv}, 2025.

\bibitem{Germinal}
L.~S. Mille-Fragoso, J.~N. Wang, C.~L. Driscoll, H.~Dai, T.~Widatalla,
  X.~Zhang, B.~L. Hie, and X.~J. Gao.
\newblock Efficient generation of epitope-targeted de novo antibodies with
  {{Germinal}}.

\bibitem{proteinhunter}
Y.~Cho, G.~Rangel, G.~Bhardwaj, and S.~Ovchinnikov.
\newblock Protein {{Hunter}}: {{Exploiting}} structure hallucination within
  diffusion for protein design.

\bibitem{halludesign}
M.~Fang, C.~Wang, J.~Shi, F.~Lian, Q.~Jin, Z.~Wang, Z.~Cui, Y.~Wang, Y.~Ke,
  Q.~Han, and L.~Cao.
\newblock {{HalluDesign}}: {{Protein Optimization}} and de {{Novo Design}} via
  {{Iterative Structure Hallucination}} and {{Sequence Design}}.

\bibitem{liu2026torchfold}
Y.~Liu, D.~Wang, Z.~Li, S.~Gao, H.~Wang, Z.~Shen, Z.~Ma, Y.~Zhang, and M.~Chen.
\newblock {{TorchFold}}: A {{PyTorch}} and {{NPU-compatible}} reimplementation
  of {{AlphaFold3}}, 2026.

\bibitem{shuaiIgLMInfillingLanguage2023}
R.~W. Shuai, J.~A. Ruffolo, and J.~J. Gray.
\newblock {{IgLM}}: {{Infilling}} language modeling for antibody sequence
  design.
\newblock {\em Cell Systems}, 14(11):979--989.e4, 2023.

\bibitem{TNP}
G.~L. Gordon, J.~Gervasio, C.~Souders, and C.~M. Deane.
\newblock The {{Therapeutic Nanobody Profiler}}: Characterising and predicting
  nanobody developability to improve therapeutic design.

\bibitem{BioPhi}
D.~Prihoda, J.~Maamary, A.~Waight, V.~Juan, L.~Fayadat-Dilman, D.~Svozil, and
  D.~A. Bitton.
\newblock {{BioPhi}}: {{A}} platform for antibody design, humanization, and
  humanness evaluation based on natural antibody repertoires and deep learning.
\newblock {\em mAbs}, 14(1):2020203, 2022.

\bibitem{rfantibody}
N.~R. Bennett, J.~L. Watson, R.~J. Ragotte, A.~J. Borst, D.~L. See, C.~Weidle,
  R.~Biswas, Y.~Yu, E.~L. Shrock, R.~Ault, P.~J.~Y. Leung, B.~Huang,
  I.~Goreshnik, J.~Tam, K.~D. Carr, B.~Singer, C.~Criswell, B.~I.~M. Wicky,
  D.~Vafeados, M.~G. Sanchez, H.~M. Kim, S.~Vázquez~Torres, S.~Chan, S.~M.
  Sun, T.~Spear, Y.~Sun, K.~O’Reilly, J.~M. Maris, N.~G. Sgourakis, R.~A.
  Melnyk, C.~C. Liu, and D.~Baker.
\newblock Atomically accurate de novo design of antibodies with
  {{RFdiffusion}}, 2024.

\bibitem{complexa}
K.~Didi, D.~Reidenbach, M.~Penner, S.~Ravichandran, M.~Nichols, E.~Swanson,
  A.~Reis, M.~Prescott, Y.~Qian, D.~Qian, J.~Yang, W.~Li, L.~Li, D.~Shonai,
  S.~Gay, B.~B. Mallik, H.~Yeung, L.~Chen, M.~A. Juantay, H.~Klein, A.~U.
  Macintyre, D.~Granata, Z.~Cao, G.~Zhou, T.~Geffner, X.~Chen, Z.~Zhang,
  T.~Zhang, K.~Gion, M.~M. Bronstein, K.~Deibler, S.~Soderling,
  A.~Khmelinskaia, F.~Hollfelder, C.~Dallago, E.~Kucukbenli, A.~Vahdat,
  P.~Ogden, and K.~Kreis.
\newblock Latent {{Generative Search}} unlocks de novo {{Design}} of {{Untapped
  Biomolecular Interactions}} at {{Scale}}.

\bibitem{rosettafoldaa}
R.~Krishna, J.~Wang, W.~Ahern, P.~Sturmfels, P.~Venkatesh, I.~Kalvet, G.~R.
  Lee, F.~S. Morey-Burrows, I.~Anishchenko, I.~R. Humphreys, R.~McHugh,
  D.~Vafeados, X.~Li, G.~A. Sutherland, A.~Hitchcock, C.~N. Hunter, A.~Kang,
  E.~Brackenbrough, A.~K. Bera, M.~Baek, F.~DiMaio, and D.~Baker.
\newblock Generalized biomolecular modeling and design with {{RoseTTAFold
  All-Atom}}.
\newblock {\em Science}, 384(6693):eadl2528, 2024.

\bibitem{AutoDock2010}
O.~Trott and A.~J. Olson.
\newblock {{AutoDock Vina}}: {{Improving}} the speed and accuracy of docking
  with a new scoring function, efficient optimization, and multithreading.
\newblock {\em J Comput Chem}, 31(2):455--461, 2010.

\bibitem{AutoDock2021}
J.~Eberhardt, D.~Santos-Martins, A.~F. Tillack, and S.~Forli.
\newblock {{AutoDock Vina}} 1.2.0: {{New Docking Methods}}, {{Expanded Force
  Field}}, and {{Python Bindings}}.
\newblock {\em J. Chem. Inf. Model.}, 61(8):3891--3898, 2021.

\end{thebibliography}

\clearpage

\appendix

\section*{Appendix}
\noindent\textsc{Contents}
\begingroup
\small
\appendixcontentsline{1}{1.5em}{6.0em}{Appendix A Methods}{app:methods}
\appendixcontentsline{2}{3.8em}{4.0em}{A.1 TorchCraft design protocol}{app:design-protocol}
\appendixcontentsline{3}{6.4em}{5.0em}{A.1.1 General optimization framework}{app:general-optimization-framework}
\appendixcontentsline{3}{6.4em}{5.0em}{A.1.2 Four-stage optimization}{app:four-stage-optimization}
\appendixcontentsline{3}{6.4em}{5.0em}{A.1.3 Target-specific stage configurations}{app:target-specific-stage-configurations}
\appendixcontentsline{3}{6.4em}{5.0em}{A.1.4 Design losses}{app:design-losses}
\appendixcontentsline{2}{3.8em}{4.0em}{A.2 Experimental-candidate design protocol}{app:experimental-candidate-design-protocol}
\appendixcontentsline{3}{6.4em}{5.0em}{A.2.1 Design task specification}{app:design-task-specification}
\appendixcontentsline{3}{6.4em}{5.0em}{A.2.2 Design experiments}{app:design-experiments}
\appendixcontentsline{4}{9.2em}{6.0em}{A.2.2.1 Minibinder candidate generation and selection}{app:minibinder-candidate-selection}
\appendixcontentsline{4}{9.2em}{6.0em}{A.2.2.2 VHH candidate generation and selection}{app:vhh-candidate-selection}
\appendixcontentsline{3}{6.4em}{5.0em}{A.2.3 Structural novelty and diversity analyses}{app:structural-novelty-diversity}
\appendixcontentsline{2}{3.8em}{4.0em}{A.3 Implementation optimizations}{app:implementation-optimizations}
\appendixcontentsline{3}{6.4em}{5.0em}{A.3.1 Multithreaded parallel conformer generation}{app:parallel-conformer-generation}
\appendixcontentsline{3}{6.4em}{5.0em}{A.3.2 Gradient rematerialization}{app:gradient-rematerialization}
\appendixcontentsline{3}{6.4em}{5.0em}{A.3.3 Caching mechanism}{app:caching-mechanism}
\appendixcontentsline{3}{6.4em}{5.0em}{A.3.4 NPU Fusion Attention}{app:npu-fusion-attention}
\appendixcontentsline{3}{6.4em}{5.0em}{A.3.5 Other optimizations}{app:other-optimizations}
\appendixcontentsline{2}{3.8em}{4.0em}{A.4 Benchmark settings}{app:benchmark-settings}
\appendixcontentsline{3}{6.4em}{5.0em}{A.4.1 Target information}{app:target-information}
\appendixcontentsline{3}{6.4em}{5.0em}{A.4.2 TorchScore}{app:torchscore}
\appendixcontentsline{3}{6.4em}{5.0em}{A.4.3 Method setup}{app:method-setup}
\appendixcontentsline{2}{3.8em}{4.0em}{A.5 Experimental validation methods}{app:experimental-validation-methods}
\appendixcontentsline{3}{6.4em}{5.0em}{A.5.1 Binder experimental methods}{app:binder-experimental-methods}
\appendixcontentsline{4}{9.2em}{6.0em}{A.5.1.1 Gene synthesis and subcloning}{app:binder-gene-synthesis}
\appendixcontentsline{4}{9.2em}{6.0em}{A.5.1.2 Cell-free protein expression}{app:binder-expression}
\appendixcontentsline{4}{9.2em}{6.0em}{A.5.1.3 Cell-free protein purification and analysis}{app:binder-purification}
\appendixcontentsline{4}{9.2em}{6.0em}{A.5.1.4 Target protein preparation}{app:binder-target-preparation}
\appendixcontentsline{4}{9.2em}{6.0em}{A.5.1.5 Binding characterization of binders}{app:binder-binding-characterization}
\appendixcontentsline{3}{6.4em}{5.0em}{A.5.2 VHH experimental methods}{app:vhh-experimental-methods}
\appendixcontentsline{4}{9.2em}{6.0em}{A.5.2.1 Framework selection}{app:vhh-framework-selection}
\appendixcontentsline{4}{9.2em}{6.0em}{A.5.2.2 Gene synthesis and subcloning}{app:vhh-gene-synthesis}
\appendixcontentsline{4}{9.2em}{6.0em}{A.5.2.3 VHH expression in CHO cells}{app:vhh-expression}
\appendixcontentsline{4}{9.2em}{6.0em}{A.5.2.4 Protein purification and analysis}{app:vhh-purification}
\appendixcontentsline{4}{9.2em}{6.0em}{A.5.2.5 Target protein preparation}{app:vhh-target-preparation}
\appendixcontentsline{4}{9.2em}{6.0em}{A.5.2.6 Binding characterization of VHHs}{app:vhh-binding-characterization}
\appendixcontentsline{1}{1.5em}{6.0em}{Appendix B Supplementary Tables}{app:supplementary-tables}
\appendixcontentsline{1}{1.5em}{6.0em}{Appendix C Supplementary Figures}{app:supplementary-figures}
\endgroup

\clearpage
\section{Methods}
\label{app:methods}

\subsection{TorchCraft design protocol}
\label{app:design-protocol}

\subsubsection{General optimization framework}
\label{app:general-optimization-framework}

TorchCraft design trajectories are configured from a target molecular context, an editable binder
sequence, and optional restraints. Target structures are supplied as templates, with crops and
hotspot definitions recorded for each task. Editable positions may span the complete binder or
selected residues within a fixed sequence framework. Task configurations also specify excluded
residue types, secondary-structure objectives, and sequence priors. This setup allows the same
optimization engine to be used across minibinders, VHHs, cyclic peptides, and ligand-binding
proteins while changing the editable positions, loss terms, molecular constraints, and optimization
schedules.

The editable sequence is parameterized by a logit matrix $\mathbf{L}$ with one row per editable
position and one column per standard amino acid. At each step, the stage-specific sequence
representation is passed through a frozen TorchFold predictor, and the weighted design objective is
differentiated with respect to $\mathbf{L}$. Only the editable sequence variables are updated.
Active objectives and their weights are specified separately for each task in Supplementary
Table~\ref{tab:design_threeline}. All designs reported in this study used pretrained AF3 weights.
The final sequence is obtained by discrete decoding, followed by optional sequence redesign only in
the explicitly designated benchmark settings.

\subsubsection{Four-stage optimization}\label{app:four-stage-optimization}
TorchCraft uses a configurable sequence-optimization schedule adapted from prior
predictor-inversion methods.\cite{boltzdesign,bindcraft} The schedule can include a soft warm-up, a
logits-to-probabilities transition, temperature annealing, and straight-through discrete
optimization. Task-specific subsets of these stages are used, as listed in Supplementary
Table~\ref{tab:design_threeline}. The sequence to be optimized is parameterized as a trainable logit
matrix $\mathbf{L}$, which is then converted into the model input sequence representation
$\mathbf{X}$ according to the optimization stage. Given $\mathbf{X}$, the frozen predictor produces
the structural state $\hat{\mathcal{S}}$, and gradients of the design objective
$\mathcal{L}(\hat{\mathcal{S}}, \mathbf{X})$ are backpropagated to update the sequence logits.
Algorithm~\ref{alg:four_stage_optimization} summarizes the conversion from sequence logits to
predictor inputs and the resulting gradient update.

\textbf{Soft warm-up (Stage 0).}
During the optional soft warm-up, sequence logits are converted to probabilities to provide
normalized inputs before subsequent optimization:
\[
  \mathbf{X} = \mathrm{Softmax}(\mathbf{L}) .
\]
The target context and active losses during warm-up are task-dependent, as specified in
Appendix~\ref{app:target-specific-stage-configurations}.

\textbf{Logits-to-soft transition (Stage 1).}
A linearly scheduled mixing coefficient $a$ is introduced to interpolate between the raw logit
representation and the softmax-normalized probabilistic representation:
\[
  \mathbf{X} = (1-a)\mathbf{L} + a\,\mathrm{Softmax}(\mathbf{L}) ,
\]
where $a$ is gradually increased over the course of this stage. This transitions the input from the
raw logit representation toward a normalized sequence representation.

\textbf{Temperature annealing (Stage 2).}
A temperature parameter $\tau$ is used to control the sharpness of the softmax distribution:
\[
  \mathbf{X} = \mathrm{Softmax}(\mathbf{L}/\tau) .
\]
Given an initial temperature $\tau_0$, $\tau$ is annealed according to
\[
  \tau = \max\left(\tau_{\min},\; \tau_0 \cdot
  \left(\frac{\tau_{\min}}{\tau_0}\right)^{t/T}\right) .
\]
As the temperature decreases, the sequence distribution sharpens from a smooth probability
distribution toward a near one-hot representation. Here $t$ is the step within the annealing stage
and $T$ is the number of steps in that stage. The temperature decreases from $\tau_0$ to
$\tau_{\min}$, sharpening each position's amino-acid distribution.

\textbf{Straight-through hard optimization (Stage 3).}
The sequence representation is explicitly converted into a one-hot form. Gradients are still
propagated through the underlying softmax probabilities. Low-temperature softmax probabilities are
computed as
\[
  \mathbf{P} = \mathrm{Softmax}(\mathbf{L}/\tau_{\min}) ,
\]
and the amino acid with the highest probability is selected to form the hard one-hot representation
$\mathbf{H} = \mathrm{OneHot}(\arg\max \mathbf{P})$. The final sequence representation used as model
input is defined as
\[
  \mathbf{X} = (\mathbf{H}-\mathbf{P})_{\mathrm{stopgrad}} + \mathbf{P} .
\]
The forward pass uses the one-hot sequence $\mathbf{H}$, while the backward pass uses the
derivative of $\mathbf{P}$ with respect to the logits. This straight-through estimator permits
additional sequence updates under a discrete forward representation.

\textbf{Final decoding.}
After the final stage, each editable position is assigned its highest-logit amino acid. Optional
sequence-redesign outputs are retained as a separate method setting.

\begin{algorithm}[H]
\DontPrintSemicolon
\caption{Four-Stage Optimization}
\label{alg:four_stage_optimization}

\SetKwInput{Input}{Input}
\SetKwInput{Output}{Output}
\Input{
    Initial sequence logits $\mathbf{L}$, frozen predictor $f_{\theta}$\;
    Design objective $\mathcal{L}$, Stages $\{s_0, s_1, s_2, s_3\}$\;
}

\Output{Optimized discrete sequence $S^{\star}$\;}

\For{$\mathrm{optimization\ step\ } t$}{
    \BlankLine
    \tcp{Stage 0: soft warm-up}
    \If{$t \in s_0$}{
        $\mathbf{X} \leftarrow \mathrm{Softmax}(\mathbf{L})$\;
    }

    \BlankLine
    \tcp{Stage 1: logits-to-soft transition}
    \ElseIf{$t \in s_1$}{
        $a \leftarrow \mathrm{LinearSchedule}(t)$\;
        $\mathbf{P} \leftarrow \mathrm{Softmax}(\mathbf{L})$\;
        $\mathbf{X} \leftarrow (1-a)\mathbf{L} + a\mathbf{P}$\;
    }

    \BlankLine
    \tcp{Stage 2: temperature annealing}
    \ElseIf{$t \in s_2$}{
        $\tau \leftarrow \mathrm{AnnealTemperature}(t)$\;
        $\mathbf{X} \leftarrow \mathrm{Softmax}(\mathbf{L}/\tau)$\;
    }

    \BlankLine
    \tcp{Stage 3: straight-through hard optimization}
    \ElseIf{$t \in s_3$}{
        $\mathbf{P} \leftarrow \mathrm{Softmax}(\mathbf{L}/\tau_{\min})$\;
        $\mathbf{H} \leftarrow \mathrm{OneHot}(\arg\max \mathbf{P})$\;
        $\mathbf{X} \leftarrow (\mathbf{H}-\mathbf{P})_{\mathrm{stopgrad}} + \mathbf{P}$\;
    }

    \BlankLine
    $\hat{\mathcal{S}} \leftarrow f_{\theta}(\mathbf{X})$\;
    $\mathbf{L} \leftarrow \mathrm{UpdateLogits}(\mathbf{L}, \nabla_{\mathbf{L}}\mathcal{L}(\hat{\mathcal{S}}, \mathbf{X}))$\;
}

\BlankLine
$S^{\star} \leftarrow \arg\max \mathrm{Softmax}(\mathbf{L})$\;
\Return{$S^{\star}$}\;

\end{algorithm}

\subsubsection{Target-specific stage configurations}\label{app:target-specific-stage-configurations}
Minibinder and VHH runs used temperature annealing followed by straight-through optimization.
Cyclic-peptide runs used all four stages. Stage lengths and learning rates are listed in
Supplementary Table~\ref{tab:design_threeline}.

Small-molecule design used an initial ligand-free warm-up followed by gradual introduction of the
interface objective. After this warm-up, the ligand was added and the complex was jointly
optimized. We write the time-dependent objective as
\[
  \mathcal{L}(t) = \mathcal{L}_{\mathrm{scaffold}} + w_{\mathrm{interface}}(t) \cdot
  \mathcal{L}_{\mathrm{interface}},
\]
where $w_{\mathrm{interface}}(t)$ increases from zero to its final value during the specified
transition stage and remains fixed thereafter. The molecular inputs, active losses, and stage
lengths are reported in Supplementary Table~\ref{tab:design_threeline}.

\subsubsection{Design losses}
\label{app:design-losses}

\textbf{Confidence losses.}
Loss terms based on the following confidence scores are constructed:

\textit{pLDDT loss.}
pLDDT (predicted Local Distance Difference Test) measures local structural confidence. This loss
increases predicted local confidence of binder residues or atoms.

\textit{pTM loss.}
pTM (predicted TM-score) measures predicted global confidence. This loss increases predicted
global confidence for the selected structure.

\textit{ipTM loss.}
ipTM (interface predicted TM-score) evaluates predicted confidence in the relative placement of
different subunits. This loss increases predicted interface confidence between chains or molecular
components.

\textit{PDE loss.}
PDE (predicted Distance Error) penalizes higher predicted distance error between residue or atom
pairs.

\textit{PAE loss.}
PAE (predicted Aligned Error) reflects uncertainty in the relative spatial placement of residue or
token pairs. This loss penalizes higher predicted alignment error.

\textit{iPAE loss.}
iPAE (interface predicted Aligned Error) focuses on predicted alignment error across chains or
molecular components at the complex interface.

\textbf{Contact loss.}
We impose soft contact constraints using the predicted distogram rather than the final inter-residue
distances in the predicted structure. For each residue pair $(i,j)$, the distogram logits are used to
estimate a probability distribution over distance bins. Given a distance threshold, all bins below
the threshold are treated as contact bins, and the contact loss is computed from the corresponding
probabilities.

In the optional binary formulation, the loss mainly penalizes insufficient total probability mass
within the contact region.
Otherwise, the probability distribution is renormalized over the contact bins, and the corresponding
cross-entropy term is computed. For each anchor residue, only a subset of partner residues with the
lowest losses is retained. The best-performing fraction of anchor residues is then selected and
averaged. This aggregation rewards a selected subset of high-probability contacts.

\textit{Intra-chain binder contact loss.}
The intra-chain binder contact loss encourages the designed binder to form a sufficient number of
internal residue--residue contacts. This biases the binder toward a more compact fold.

\textit{Inter-chain target--binder contact loss.}
The inter-chain target--binder contact loss encourages the designed binder to form a sufficient
number of cross-chain contacts with the target. This strengthens interface formation between the two
chains. If hotspot residues are provided, the loss prioritizes contacts between the binder and the
specified target sites. If the designable positions on the binder side are further restricted,
interface contacts are mainly encouraged between those designable binder residues and the target or
hotspot region.

\textbf{$\alpha$-Helix loss.}
Inspired by BoltzDesign,\cite{boltzdesign} an $\alpha$-Helix loss is introduced to explicitly
encourage, or suppress when assigned a negative weight, the formation of $\alpha$-helical
conformations. It uses contact probabilities derived from the predicted distance distribution as a
differentiable signal, and rewards residue pairs $i$ and $i+3$ within $6~\text{\AA}$, as residues
separated by three positions along the sequence are typically in close spatial proximity in
$\alpha$-helices. Formally, the $\alpha$-Helix loss is equivalent to minimizing the binary contact
negative log-likelihood over selected $(i,i+3)$ pairs:
\[
  \mathcal{L}_{\alpha}
  = -\frac{1}{|H|}\sum_{(i,j)\in H}\log(p^{\alpha}_{ij}+\epsilon),
\]
where the candidate helical contact set is defined within the binder chain as
\[
  H = \{(i,j)\mid |i-j|=3\}.
\]
Given the predicted pairwise distance distribution $P_{ij}(d)$, the confidence that a residue pair
forms a local contact is defined as the probability mass below the distance threshold of
$6~\text{\AA}$:
\[
  p^{\alpha}_{ij} = \sum_{d < 6~\text{\AA}} P_{ij}(d).
\]

\textbf{$\beta$-structure loss.}
We introduced a $\beta$-structure loss to modulate the $\beta$-structure propensity of the binder,
similar to Germinal.\cite{Germinal} Together with the $\alpha$-helix loss, this provides more
flexible control over the secondary-structure composition of designed samples. The $\beta$-structure
loss consists of a $\beta$-strand term and a $\beta$-sheet term:
\[
  \mathcal{L}_{\beta}
  = \mathcal{L}_{\beta\text{-strand}} + \mathcal{L}_{\beta\text{-sheet}}.
\]
For local $\beta$-strand geometry, the probability of a residue-pair distance falling within the
interval $[9.75,11.5)~\text{\AA}$ is defined as
\[
  p^{\beta\text{-strand}}_{ij}
  = \sum_{d\in[9.75,11.5)~\text{\AA}} P_{ij}(d).
\]
Let $B$ denote the candidate $\beta$-design positions, define their neighboring position set as
\[
  N = \{i-1,i+1\mid i\in B\},
\]
and the local $\beta$-strand candidate pairs as
\[
  S = \{(i,j)\mid i,j\in N,\; j=i+3\}.
\]
Among these candidates, we select the top-$K$ pairs with the largest per-pair losses, focusing
optimization on the local $\beta$-strand constraints that are currently least satisfied, where
\[
  K = \left\lfloor \frac{|B|}{3} \right\rfloor.
\]
The $\beta$-strand loss is then given by
\[
  \mathcal{L}_{\beta\text{-strand}}
  = -\frac{1}{K}\sum_{(i,j)\in S_{\mathrm{top}\text{-}K}}
  \log(p^{\beta\text{-strand}}_{ij}+\epsilon).
\]
For $\beta$-sheet formation, a shorter contact-distance interval,
$[4.4,6.0)~\text{\AA}$, is used to characterize spatial contacts between two $\beta$-strands:
\[
  p^{\beta\text{-sheet}}_{ij}
  = \sum_{d\in[4.4,6.0)~\text{\AA}} P_{ij}(d).
\]
The $\beta$-sheet term enumerates a set of antiparallel $\beta$-sheet pairing patterns. For each
candidate $\beta$-region $R$, we construct multiple reverse-diagonal pairing masks, each
corresponding to a possible strand--strand pairing configuration. We denote the resulting candidate
pairing sets as
\[
  A_R = \{A_1,A_2,\cdots,A_m\},
\]
where each $A$ is a set of residue pairs,
$A\subseteq\{(i,j)\mid i,j\in\text{binder}\}$. For a given candidate pairing pattern $A$, the
average $\beta$-sheet contact negative log-likelihood is defined as
\[
  l(A) = -\frac{1}{|A|}\sum_{(i,j)\in A}
  \log(p^{\beta\text{-sheet}}_{ij}+\epsilon).
\]
For each $\beta$-region, we take the maximum loss over all candidate pairing patterns:
\[
  l_{\beta\text{-sheet}}(R) = \max_{A\in A_R} l(A).
\]
The overall $\beta$-sheet loss is therefore
\[
  \mathcal{L}_{\beta\text{-sheet}}
  = \sum_R \max_{A\in A_R}
  \left[-\frac{1}{|A|}\sum_{(i,j)\in A}
  \log(p^{\beta\text{-sheet}}_{ij}+\epsilon)\right].
\]
Full details are provided in Algorithm~\ref{alg:beta_loss}.

\begin{algorithm}[H]
\DontPrintSemicolon
\caption{Beta Loss}
\label{alg:beta_loss}

\SetKwInput{Input}{Input}
\SetKwInput{Output}{Output}
\Input{Predicted distance distribution $P_{ij}(d)$; candidate $\beta$ design positions $B$; candidate $\beta$ regions $\mathcal{R}$; numerical constant $\epsilon$\;}

\Output{$\beta$ structural loss $\mathcal{L}_{\beta}$\;}

\BlankLine
\tcp{$\beta$-strand loss}
$\mathcal{L}_{\beta\text{-strand}} \leftarrow 0$\;
$N \leftarrow \{i-1, i+1 \mid i \in B\}$\;
$S \leftarrow \{(i,j) \mid i,j \in N,\; j=i+3\}$\;
$K \leftarrow \left\lfloor |B|/3 \right\rfloor$\;

\For{$(i,j) \in S$}{
    $p^{\beta\text{-strand}}_{ij}
    \leftarrow \sum_{d \in [9.75, 11.5)\,\text{\AA}} P_{ij}(d)$\;
    $\ell^{\beta\text{-strand}}_{ij}
    \leftarrow -\log\left(p^{\beta\text{-strand}}_{ij}+\epsilon\right)$\;
}

$S_{\mathrm{top}\text{-}K}
\leftarrow \mathrm{TopK}\left(S,\; \ell^{\beta\text{-strand}}_{ij},\; K\right)$\;

$\mathcal{L}_{\beta\text{-strand}}
\leftarrow
-\frac{1}{K}
\sum_{(i,j)\in S_{\mathrm{top}\text{-}K}}
\log\left(p^{\beta\text{-strand}}_{ij}+\epsilon\right)$\;

\BlankLine
\tcp{$\beta$-sheet loss}
$\mathcal{L}_{\beta\text{-sheet}} \leftarrow 0$\;

\For{$R \in \mathcal{R}$}{
    Construct antiparallel pairing masks
    $A_R=\{A_1,A_2,\cdots,A_m\}$ for $R$\;

    \For{$A \in A_R$}{
        \For{$(i,j) \in A$}{
            $p^{\beta\text{-sheet}}_{ij}
            \leftarrow
            \sum_{d \in [4.4, 6.0)\,\text{\AA}} P_{ij}(d)$\;
        }

        $l(A)
        \leftarrow
        -\frac{1}{|A|}
        \sum_{(i,j)\in A}
        \log\left(p^{\beta\text{-sheet}}_{ij}+\epsilon\right)$\;
    }

    $l_{\beta\text{-sheet}}(R)
    \leftarrow
    \max_{A\in A_R} l(A)$\;

    $\mathcal{L}_{\beta\text{-sheet}}
    \leftarrow
    \mathcal{L}_{\beta\text{-sheet}}
    +
    l_{\beta\text{-sheet}}(R)$\;
}

\BlankLine
$\mathcal{L}_{\beta}
\leftarrow
\mathcal{L}_{\beta\text{-strand}}
+
\mathcal{L}_{\beta\text{-sheet}}$\;

\Return{$\mathcal{L}_{\beta}$}\;

\end{algorithm}

\textbf{Paratope loss.}
Inspired by Germinal,\cite{Germinal} a distogram-based paratope contact loss is introduced to
restrict the designed binding interface to predefined CDR or paratope regions. The goal is to
strengthen predicted contacts between CDR residues and the target epitope or hotspot residues, while
discouraging framework--target contacts.

We compute three contact losses: the CDR--hotspot contact loss
$\mathcal{L}_{\mathrm{CDR}\text{-}\mathrm{Hotspot}}$, the CDR--interface contact loss
$\mathcal{L}_{\mathrm{CDR}\text{-}\mathrm{Interface}}$, and the framework--target contact loss
$\mathcal{L}_{\mathrm{framework}}$. Since a smaller contact loss indicates stronger predicted
contact, minimizing $\mathcal{L}_{\mathrm{CDR}\text{-}\mathrm{Hotspot}}$ and
$\mathcal{L}_{\mathrm{CDR}\text{-}\mathrm{Interface}}$ jointly promotes interface formation through
the CDR regions. In contrast, contacts between the framework and the target should remain weak. We
therefore place $\mathcal{L}_{\mathrm{framework}}$ in the denominator as a suppression term:
\[
  \mathcal{L}_{\mathrm{paratope}}
  =
  \frac{
    \mathcal{L}_{\mathrm{CDR}\text{-}\mathrm{Hotspot}}
    \cdot
    \mathcal{L}_{\mathrm{CDR}\text{-}\mathrm{Interface}}
  }{
    \operatorname{ReLU}\left(\mathcal{L}_{\mathrm{framework}}-\lambda\right)+\epsilon
  }.
\]
The denominator is lower-bounded by $\epsilon$ when
$\mathcal{L}_{\mathrm{framework}}-\lambda$ is negative or close to zero.

\textbf{Distogram-based compactness loss.}
To penalize overly extended conformations during sequence optimization, we constructed a
differentiable radius-of-gyration (ROG) proxy using the predicted distogram probability
distribution. For a binder chain containing $N$ residues, the squared radius of gyration of an
equally weighted point cloud can be expressed as the average over all pairwise Euclidean distances:
\[
  R_g^2 = \frac{1}{2N^2}\sum_i\sum_j \lVert \mathbf{r}_i-\mathbf{r}_j\rVert^2,
\]
where $\mathbf{r}_i$ denotes the spatial position of residue $i$. For each residue pair $(i,j)$, let
$b_k$ denote the $k$-th distogram bin, $c_k$ its representative distance, taken as the bin center,
and $p_{ij}(b_k)$ the corresponding predicted probability. The expected squared distance for this
residue pair is approximated as
\[
  \mathbb{E}[d_{ij}^2] = \sum_k p_{ij}(b_k)c_k^2.
\]
A distogram-based approximation of $R_g^2$ is then obtained by averaging the expected squared
distances over all residue pairs in the binder:
\[
  R_g^2 = \frac{1}{2N^2}\sum_i\sum_j \mathbb{E}[d_{ij}^2].
\]
We further define the radius of gyration as
\[
  R_g = \sqrt{R_g^2+\epsilon}.
\]
To impose a length-dependent compactness upper bound, we define the target radius of gyration as
\[
  R_{g,\mathrm{target}} = \alpha N^{\beta}+\gamma.
\]
The final ROG regularization term is defined as
\[
  \mathcal{L}_{\mathrm{rog}}
  = \operatorname{ReLU}\left(R_g - R_{g,\mathrm{target}}\right).
\]

\textbf{Isotropy loss.}
A radius-of-gyration penalty controls overall size but does not uniquely specify shape. Elongated
rods, thin disks, and spherical conformations can exhibit similar values of $R_g$. To bias the
binder toward more similar extents along its principal axes, we additionally defined an isotropy
objective from a distogram-derived distance matrix. From the distogram, we first compute the
expected squared distance for each residue pair and assemble the intra-binder
squared distance matrix:
\[
  \mathbf{D}^{(2)}_{ij} = \mathbb{E}[d_{ij}^2].
\]
The matrix is then symmetrized as
\[
  \mathbf{D}^{(2)} \leftarrow
  \frac{1}{2}\left(\mathbf{D}^{(2)}+\mathbf{D}^{(2)T}\right),
\]
and its diagonal entries are set to zero:
\[
  D^{(2)}_{ii}=0.
\]
We next recover a centered Gram matrix from the squared distance matrix using the standard
double-centering formula:
\[
  \mathbf{G} = -\frac{1}{2}\mathbf{J}\mathbf{D}^{(2)}\mathbf{J},
\]
where
\[
  \mathbf{J} = \mathbf{I}-\frac{1}{N}\mathbf{1}\mathbf{1}^{T}.
\]
When $\mathbf{D}^{(2)}$ is derived from real three-dimensional coordinates, $\mathbf{G}$ is
equivalent to the inner-product matrix of the centered coordinate matrix,
$\mathbf{G}=\mathbf{X}\mathbf{X}^{T}$. The eigenvalues of $\mathbf{G}$ reflect the variance scale of
the point cloud along its principal axes. We take the three largest non-negative eigenvalues,
$\lambda_1\geq\lambda_2\geq\lambda_3$, which correspond to the extent of the structure along the
three principal axes, and penalize deviations among them:
\[
  \mathcal{L}_{\mathrm{isotro}}
  =
  \frac{
  (\lambda_1-\lambda_2)^2
  +
  (\lambda_2-\lambda_3)^2
  +
  (\lambda_1-\lambda_3)^2
  }{
  (\lambda_1+\lambda_2+\lambda_3)^2+\epsilon
  }.
\]
For a near-spherical conformation, the three principal extents should be comparable, such that
$\lambda_1\approx\lambda_2\approx\lambda_3$.

\textbf{Amino-acid composition loss.}
An aa-type loss regularizes the global residue distribution of the optimized chain. A reference
composition is estimated from ProteinMPNN-redesigned samples, and the current sequence probabilities
are averaged over editable positions and matched to this reference by cross-entropy. This term
constrains global residue usage and is evaluated separately from position-specific structural and
language-model objectives.
Let $\mathcal{E}$ denote the editable residue positions, and let
$\mathbf{P}=\mathrm{Softmax}(\mathbf{L})$ be the amino-acid probability matrix obtained from the
current sequence logits. The empirical amino-acid composition of the optimized region is
\[
  \bar{p}_a
  =
  \frac{1}{|\mathcal{E}|}
  \sum_{i\in\mathcal{E}} P_{i,a},
  \qquad a\in\mathcal{A},
\]
where $\mathcal{A}$ is the 20-amino-acid alphabet. Given a ProteinMPNN-derived reference composition
$q_a$, the amino-acid composition loss is
\[
  \mathcal{L}_{\mathrm{AA}}
  =
  -\sum_{a\in\mathcal{A}} q_a \log\left(\bar{p}_a+\epsilon\right).
\]

\textbf{CDR-restricted IgLM loss.}
In nanobody CDR design, the CDR sequences to be optimized are represented as a continuous logit
matrix $\mathbf{L}_{\mathrm{CDR}}$. Inspired by Germinal,\cite{Germinal} we adopt a decoupled
dual-path strategy to incorporate antibody language-model priors while preserving differentiable
sequence optimization.\cite{shuaiIgLMInfillingLanguage2023}

The first path constructs the discrete context for IgLM. A pseudo sequence is obtained by taking
the $\arg\max$ of the current logits; equivalently, of a temperature-scaled softmax of those
logits. For fixed logits, this discrete context does not depend on the softmax temperature.
For each CDR, we then construct an infilling input independently: the CDR region being
evaluated is replaced with mask tokens, while the framework regions and the remaining CDRs are
provided as context to the frozen IgLM. Under this local context, IgLM predicts the conditional amino
acid distribution at each masked position, yielding a context-dependent target distribution
$\hat{\mathbf{P}}_{\mathrm{IgLM}}$.

The second path is used for differentiable gradient propagation. The original logits are directly
passed through a softmax with temperature 1 to obtain the continuous predicted distribution
$\mathbf{P}_{\mathrm{pred}}$. The local conditional distribution predicted by IgLM is then used as
the teacher distribution, and a cross-entropy loss is computed over the CDR region. Because the loss
is backpropagated only through the continuous $\mathbf{P}_{\mathrm{pred}}$ path, gradients can act
stably on the original logits, while the discrete pseudo sequence is used solely to provide context
for IgLM. This avoids the gradient interruption introduced by $\arg\max$ discretization, while still
using the antibody sequence prior learned by IgLM to guide CDR optimization.

We evaluated the sequence-prior objective separately over the three editable CDRs, allowing their
contributions to be weighted independently. This differs from a full-sequence objective in both the
positions scored and the conditioning context
(Algorithm~\ref{alg:iglm_cdr}).

\begin{algorithm}[H] 
  \DontPrintSemicolon 
  \caption{CDR-restricted IgLM Loss}
  \label{alg:iglm_cdr}
  
  \SetKwInput{Input}{Input} 
  \SetKwInput{Output}{Output} 
  
  \Input{ 
      Sequence logits $\mathbf{L}_{\mathrm{CDR}}$\; 
      Fixed framework sequence $S_{\mathrm{fw}}$\; 
      CDR regions $\mathcal{R}=\{R_1,\cdots,R_m\}$\; 
      CDR weights $\{w_i\}_{i=1}^{m}$\; 
  }
  
  \Output{IgLM infilling loss $\mathcal{L}_{\mathrm{IgLM}}$\;}
  
  \BlankLine 
  \tcp{Path 1: Discrete target for IgLM Context} 
  $\tau_{step} \leftarrow \text{ExponentialDecay}(\tau_{start}, \tau_{min}, \text{step})$\;
  $\mathbf{P}_{hard} \leftarrow \text{Softmax}(\mathbf{L}_{\mathrm{CDR}} / \tau_{step})$\; 
  $S_{pseudo} \leftarrow \text{Argmax}(\mathbf{P}_{hard})$ \tcp*{Pseudo-one-hot sequence} 
  
  \BlankLine 
  \tcp{Path 2: Differentiable prediction for gradient flow} 
  $\mathbf{P}_{\mathrm{pred}} \leftarrow \mathrm{Softmax}(\mathbf{L}_{\mathrm{CDR}})$\;
  
  \BlankLine 
  $S \leftarrow \text{combine } S_{\mathrm{fw}} \text{ and } S_{pseudo}$\;
  $\mathcal{L}_{\mathrm{IgLM}} \leftarrow 0$\;
  
  \For{each CDR region $R_i \in \mathcal{R}$}{ 
      \tcp{Define Context for $R_i$ using other discrete CDRs}
      $\mathit{Context}_i \leftarrow \{S_{\mathrm{fw}}\} \cup \{ S_{pseudo}[R_j] \text{ for } j \neq i \}$\;
  
      \tcp{Construct infilling prompt with $R_i$ masked}
      $\hat{\mathbf{P}}_{\mathrm{IgLM},R_i} \leftarrow f_{\theta}^{\mathrm{IgLM}}([\mathtt{MASK}] \mid \mathit{Context}_i)$\;
  
      $\mathcal{L}_i \leftarrow \mathrm{CrossEntropy}\left( \hat{\mathbf{P}}_{\mathrm{pred},R_i}, \mathbf{P}_{\mathrm{iglm},R_i} \right)$\;
  
      $\mathcal{L}_{\mathrm{IgLM}} \leftarrow \mathcal{L}_{\mathrm{IgLM}} + w_i \mathcal{L}_i$\;
  }
  
  \BlankLine 
  \Return{$\mathcal{L}_{\mathrm{IgLM}}$} \tcp*{Update via Path 2} 
  
  \end{algorithm}

\subsection{Experimental-candidate design protocol}
\label{app:experimental-candidate-design-protocol}

\subsubsection{Design task specification}
\label{app:design-task-specification}

\textbf{Minibinder design.}
Experimental minibinders were designed against the four protein targets reported in
Figure~\ref{fig:2}a using pretrained AF3 weights through TorchFold. For each target, TorchCraft
generated 3,000 candidate molecules using the
target crops, hotspot definitions, and binder-length ranges listed in
Table~\ref{tab:minibinder_validation_settings}. The design task allowed the full binder sequence to
be optimized while keeping the target template fixed. The candidates selected for experimental
validation were raw TorchCraft outputs and were not subjected to post hoc MPNN sequence redesign.

\textbf{VHH design.}
Experimental VHHs were designed against the four protein targets reported in Figure~\ref{fig:3}a
using pretrained AF3 weights through TorchFold.
The scaffold framework, CDR length ranges, target crops, and hotspot definitions followed the
settings in Table~\ref{tab:vhh_validation_settings}. For VHH designs, TorchCraft optimized the CDR
regions under the fixed framework context, with the interface objective restricted toward CDR-driven
target recognition. The candidates selected for experimental validation were raw TorchCraft outputs
under this framework-conditioned setting, without post hoc MPNN sequence redesign.

\subsubsection{Design experiments}
\label{app:design-experiments}

\paragraph{Minibinder candidate generation and selection}\label{app:minibinder-candidate-selection}
For each minibinder target, 3,000 TorchCraft candidates were sampled. After generation, candidates
were filtered by radius of gyration (ROG $<17$), binder pTM ($>0.7$), and TorchCraft ipTM
($>0.7$). Passing candidates were then ranked using Rosetta Interface Analyzer. Candidates were
selected for experimental validation using the composite score,
\[
  \mathrm{score}_{\mathrm{composite}}
  = 100 \times \mathrm{ipTM}_{\mathrm{TorchCraft}}
  - \mathrm{score}_{\mathrm{Rosetta\ interface}} .
\]
Higher composite scores were prioritized for wet-lab validation.

\paragraph{VHH candidate generation and selection}\label{app:vhh-candidate-selection}
For each VHH target, 12,500 TorchCraft candidates were generated. Candidates were filtered using
CDR interface rate ($>0.6$), CDR3 hotspot
coverage rate ($>0$), and TorchCraft ipTM ($>0.7$). Passing candidates were ranked using Rosetta
Interface Analyzer with the same composite score used for minibinders,
\[
  \mathrm{score}_{\mathrm{composite}}
  = 100 \times \mathrm{ipTM}_{\mathrm{TorchCraft}}
  - \mathrm{score}_{\mathrm{Rosetta\ interface}} .
\]
This ranking favors designs that combine high TorchCraft interface confidence with favorable
Rosetta interface scores, and the top-ranked designs were selected for experimental validation.

\subsubsection{Structural novelty and diversity analyses}
\label{app:structural-novelty-diversity}

\textbf{Structural novelty.}
We assessed the divergence of predicted binding configurations from interactions identified around
similar target chains in the PDB. Target-similar chains were identified by taking the union of a
sequence search
(\texttt{mmseqs easy-search -s 7.5 -{}-max-seqs 1000}) and a structure search
(\texttt{foldseek easy-search -s 7.5 -{}-alignment-type 1 -{}-max-seqs 1000}) of the target
chain against the PDB. For each target-similar chain, all protein chains with one or more
heavy-atom contacts ($<5~\text{\AA}$) were taken as its binders. After superposing the design
target onto the target-similar chain with CEAlign, structural divergence was measured as the
nearest-neighbour binder C$\alpha$ RMSD; for each design, we report the minimum target-aligned
binder C$\alpha$ RMSD over all interactions with target-similar chains.

\textbf{Structural diversity.}
We clustered predicted binding configurations using average-linkage hierarchical clustering of the
target-aligned binder C$\alpha$ RMSD matrix. Using the same structural distance applied between
pairs of designs, we computed the pairwise distance matrix
$D$ over all designs in a cohort, built a dendrogram by UPGMA average-linkage clustering
(\texttt{scipy.cluster.hierarchy.linkage(squareform(D), method="average")}), and obtained the
number of clusters at a given RMSD threshold by cutting the dendrogram at that linkage
threshold
(\texttt{scipy.cluster.hierarchy.fcluster}, \texttt{criterion="distance"}). The number of clusters
as a function of the RMSD threshold is reported per cohort.

\subsection{Implementation optimizations}
\label{app:implementation-optimizations}

\subsubsection{Multithreaded parallel conformer generation}
\label{app:parallel-conformer-generation}

CPU multithreading is implemented for residue-level parallel computing. Parallel determinism is
supported through pre-extraction of random-state snapshots, and each thread maintains an
independent RandomState instance to avoid race conditions during generation of protein reference
conformations.

\subsubsection{Gradient rematerialization}
\label{app:gradient-rematerialization}

The memory bottleneck of TorchCraft stems from the large number of
activations, complex computational graphs, and high-complexity structure-related intermediate
tensors. We used gradient checkpointing to reduce activation memory by recomputing selected
intermediate tensors during backpropagation. For TorchCraft, a reentrant gradient-rematerialization
implementation was used together with native \texttt{torch.utils.checkpoint} for selected modules.

Furthermore, the Template and MSA modules exhibit no significant computational-graph nesting that
leads to memory accumulation. For these modules, non-reentrant gradient checkpointing from native
\texttt{torch.utils.checkpoint} was used.

\subsubsection{Caching mechanism}
\label{app:caching-mechanism}

Caching was restricted according to feature dependence. Features that remained unchanged were
reused, while sequence-dependent conditioning was recomputed when the sequence representation
changed. Within a single diffusion prediction, conditioning could be reused across denoising steps.

Before diffusion begins, the model runs the trunk network, such as Pairformer, once on the current
sequence representation and target context, producing \texttt{single\_cond} and
\texttt{pair\_cond}. These conditioning tensors are reused across denoising steps within the same
prediction and refreshed at the next design update. 

\subsubsection{NPU Fusion Attention}
\label{app:npu-fusion-attention}

The NPU fusion-attention operator was modified to support diverse input layouts, reducing data
movement and transpose operations. Fused attention reduced runtime for several larger sequence
settings in the reported NPU benchmark, while its effect depended on sequence length and batch size
(Supplementary Table~\ref{tab:torchdesign_speed}).

\subsubsection{Other optimizations}
\label{app:other-optimizations}

Operator substitutions and memory-layout adjustments were used on Ascend NPU, including reduced
random memory access and contiguous allocations where possible. On GPU, optional substitution of
equivariant kernels from cuequivariance was evaluated. Runtime benchmarks for NPU with and without
fused attention, and for GPU with and without cuequivariance, are summarized in
Supplementary Table~\ref{tab:torchdesign_speed}.

\subsection{Benchmark Settings}
\label{app:benchmark-settings}

\subsubsection{Target information}
\label{app:target-information}

The target structures, crop regions, and hotspot residues used in the wet-lab validation campaigns
and computational benchmarks are summarized in Supplementary Tables~\ref{tab:minibinder_validation_settings}--\ref{tab:cyclic_peptide_benchmark_targets}.

\subsubsection{TorchScore}
\label{app:torchscore}

TorchScore is a score-only loading of the TorchFold implementation.\cite{liu2026torchfold} Given
input coordinates, it bypasses the diffusion-based structure module and evaluates the complex
through the confidence pathway, returning predictor-derived scores including pLDDT, pTM, and ipTM.
No separate scoring network is trained. In this study, TorchScore used pretrained AF3 weights, the
same checkpoint used for design. Designed sequences that were rebuilt with FlowPacker were scored
from the resulting coordinates; they were not refolded from sequence by TorchScore.

\subsubsection{Method setup}
\label{app:method-setup}

\paragraph{Minibinder benchmark.}
For de novo minibinder design, we compared BoltzGen, RFdiffusion, and TorchCraft. For each method,
at each benchmark binder length and for each target, 100 designs were first generated using the
default settings of the corresponding method. For each BoltzGen and RFdiffusion design, we then
redesigned the full binder sequence with ProteinMPNN without fixing any residue, sampling eight
redesigned sequences per generated backbone. Each redesigned sequence was rebuilt into an all-atom
side-chain model with FlowPacker. The eight rebuilt models were scored with TorchScore, and the
redesign with the highest pTM + ipTM was selected as the final structure and score for that original
design. For TorchCraft, we evaluated raw and MPNN-polished settings separately. In the raw setting,
which corresponds to the default TorchCraft core procedure, outputs were scored directly with
TorchScore without sequence redesign. In the MPNN-polished setting, following the BindCraft-style
protocol, TorchCraft-designed interface residues were fixed and all remaining positions were
redesigned eight times with ProteinMPNN; FlowPacker side-chain rebuilding and TorchScore-based
selection were then performed exactly as for BoltzGen and RFdiffusion. A minibinder design was
counted as passing the computational criterion when both TorchScore pTM and TorchScore ipTM exceeded 0.7.

\paragraph{VHH benchmark.}
For VHH design, we compared BoltzGen, RFantibody, Germinal, and TorchCraft. For each method,
at each benchmark framework and for each target, 100 designs were first generated using the default
settings of the corresponding method. For BoltzGen and RFantibody, the full designed sequence was
redesigned eight times with AbMPNN without fixing any residue. For Germinal, we followed the original
pipeline by fixing CDR residues that contacted the target interface and applying AbMPNN redesign only
to residues outside the CDR regions. Each redesigned sequence was rebuilt with FlowPacker to recover
an all-atom side-chain structure. The eight rebuilt models were scored with TorchScore, and the
redesign with the highest ipTM was selected as the final structure and score for that original
design. For TorchCraft, we again evaluated raw and MPNN-polished settings separately. Raw
TorchCraft outputs were scored directly with TorchScore without sequence redesign. In the
MPNN-polished setting, interface CDR residues designed by TorchCraft were fixed and all other
positions were redesigned eight times with AbMPNN. The subsequent FlowPacker rebuilding and
TorchScore-based selection were identical to those used for BoltzGen, RFantibody, and Germinal. A VHH design passed the computational criterion only when it satisfied both a structural-confidence
criterion and an expression proxy: TorchScore ipTM had to exceed 0.7, and ESM2-150M perplexity had to
be below 6.31. The
perplexity threshold was obtained from a retrospective nanobody-expression analysis, in which public
expression measurements were combined with accumulated TorchCraft expression data. Designs with
total expression of at least 0.08 mg were classified as high-expression cases, ESM2-150M perplexity
was swept as a one-dimensional classifier, and the operating threshold was selected by maximizing the
Matthews correlation coefficient.

\paragraph{Cyclic-peptide benchmark.}
For cyclic-peptide design, we compared BoltzGen, RFpeptides, and TorchCraft. In the head-to-tail
cyclization benchmark, each method generated 100 designs per target using its default settings. For
BoltzGen and RFpeptides, the full peptide sequence of each generated structure was redesigned eight
times with ProteinMPNN without fixing any residue. Each redesigned sequence was rebuilt with
FlowPacker, scored with TorchScore, and the redesign with the highest pLDDT + 100 $\times$ ipTM was
selected as the final structure and score for that original design. For head-to-tail TorchCraft
outputs, we evaluated raw designs scored directly with TorchScore and an optional
ProteinMPNN-polished variant in which all peptide positions were redesigned eight times; the
subsequent FlowPacker and TorchScore selection procedure was identical to that used for BoltzGen and
RFpeptides. TorchCraft was also evaluated in a disulfide-linked cyclization setting. For this setting,
raw outputs were scored directly, or all positions except the two cysteines were redesigned eight
times with ProteinMPNN, followed by the same FlowPacker rebuilding and TorchScore selection protocol
used for the head-to-tail setting. A cyclic-peptide design passed the computational criterion only if the predicted
structure formed the intended cycle without chain breaks and satisfied TorchScore pLDDT $>70$ and
ipTM $>0.5$.

\paragraph{Small-molecule binder benchmark.}

For small-molecule binder design, we compared BoltzDesign and TorchCraft. For each method
and each target, 500 designs were generated using the default settings of the corresponding method.
BoltzDesign and the raw TorchCraft setting were evaluated without sequence redesign. The
score distributions of the resulting designs were assessed with AutoDock Vina and Boltz-2.
For TorchCraft, we additionally reported a TorchScore-based computational pass rate, where a design
passed the computational criterion when TorchScore pLDDT exceeded 70 and TorchScore ipTM exceeded 0.7. We also tested a
TorchCraft redesign variant in which ligand-interface residues were fixed, all remaining positions
were redesigned eight times with LigandMPNN, and the redesigned sequences were refolded with
TorchFold. The refolded structure with the highest pTM was selected as the redesigned structure and
was then evaluated with AutoDock Vina and Boltz-2.

\subsection{Experimental validation methods}
\label{app:experimental-validation-methods}

\subsubsection{Binder experimental methods}
\label{app:binder-experimental-methods}

The following procedures were used for the wet-lab validation of binder designs reported in
Figure~\ref{fig:2}a.

\paragraph{Gene synthesis and subcloning}\label{app:binder-gene-synthesis}
The DNA sequences encoding the designed minibinders were codon-optimized and synthesized by
GenScript Biotechnology Co., Ltd. (Nanjing, China). The synthesized gene was subcloned into the
pIVEX vector for protein expression in a cell-free system. Subcloning into the pIVEX vector yielded
the following product: MSG-[protein]-GSGSHHWGSTHHHHHH, containing a C-terminal SNAC cleavage tag
and a $6\times$His affinity tag.

\paragraph{Cell-free protein expression}\label{app:binder-expression}
Cell-free protein synthesis reactions were assembled by combining S30 cell lysate, synthesis
buffer, required enzymes, and plasmid DNA in a 24-deep-well plate. Each reaction had a final volume
of 5 mL and was incubated at $25\,^{\circ}\mathrm{C}$ for 16 hours. The reaction
mixture was then centrifuged at 4000 rpm for 5 minutes and the supernatant was collected for
purification.

\paragraph{Cell-free protein purification and analysis}\label{app:binder-purification}
Designed minibinders were obtained by Ni-column purification using Hamilton instruments (Hamilton,
MICRO LAB Starlet). The purified protein was then dialyzed in the desired buffer (either PBS,
pH 7.4 or 50 mM Tris, 150 mM NaCl, 10\% glycerol, pH 8.0). Before aliquoting and storage, the
protein was filtered through a 0.22 $\mu$m filter. Protein concentration was determined by the
A280 protein assay using bovine serum albumin (BSA) as the standard. Protein purity was verified by
standard SDS-PAGE electrophoresis. For SDS-PAGE analysis: samples were mixed with $5\times$
reducing loading buffer (300 mM Tris-HCl pH 6.8, 10\% SDS, 30\% glycerol, 0.5\% bromophenol blue,
and 250 mM DTT). Proteins were separated on a 12\% homogeneous SDS-PAGE gel (GenScript, Cat. No.
M00668) and visualized accordingly.

\paragraph{Target protein preparation}\label{app:binder-target-preparation}
The four minibinder campaigns used the Fc-fusion target proteins listed in
Table~\ref{tab:binder_targets}. His-tagged designed binders were immobilized for BLI measurements.

\paragraph{Binding characterization of binders}\label{app:binder-binding-characterization}
Initial screening of binder designs for target protein binding was performed using Biolayer
Interferometry (BLI, Gator Bio). For target proteins without the His-tag (IFNAR2, PD-L1, PDGFR$\beta$,
and IL-7R$\alpha$), His-tagged binders were prepared at 5 $\mu$g/mL in PBSTA buffer (10 mM
Na$_2$HPO$_4\cdot$12H$_2$O, 2 mM KH$_2$PO$_4$, 137 mM NaCl, 2.7 mM KCl, 0.05\% Tween-20, 0.1\%
BSA, pH 7.4) for loading. Target proteins were diluted to 1000 nM in PBSTA buffer. Anti-His
sensors (Gator Bio, Cat. No. 20-5066) were saturated with His-tagged binders. The BLI assay
sequence was as follows: 60 s baseline, 120 s ligand loading, 60 s post-loading baseline, 120 s
association phase, 300 s dissociation phase, and 30 s regeneration. Baseline-subtracted signals (in
nanometers) were calculated to prioritize binders for further characterization.

To determine apparent affinities of selected binder designs: His-tagged binders (5 $\mu$g/mL in PBSTA)
were immobilized onto Anti-His sensors for 120 s. Serial dilutions of target proteins were prepared
as follows: IFNAR2 (4000 nM to 62.5 nM or 125 nM to 1.95 nM), PDGFR$\beta$ (100 nM to 3.125 nM),
PD-L1 (4000 nM to 62.5 nM), and IL-7R$\alpha$ (500 nM to 31.25 nM). Diluted target proteins were
allowed to associate with the immobilized ligand for 120 s, followed by dissociation in PBSTA buffer
for 300 s. After background subtraction
using buffer-only (PBSTA) control curves, apparent dissociation constants ($K_D$) were determined using the
1:1 binding model in the Results \& Analysis curve-fitting module.

\subsubsection{VHH experimental methods}
\label{app:vhh-experimental-methods}

The following procedures were used for the wet-lab validation of VHH designs reported in
Figure~\ref{fig:3}a.

\paragraph{Framework selection}\label{app:vhh-framework-selection}
VHH frameworks were randomly selected from 7EOW (caplacizumab), 8Z8V (ozoralizumab), 5JDS
(envafolimab), 7XL0 (vobarilizumab), and 8COH (gefurulimab).

\paragraph{Gene synthesis and subcloning}\label{app:vhh-gene-synthesis}
VHH DNA sequences were codon-optimized and synthesized by GenScript Biotechnology Co., Ltd.
(Nanjing, China). The synthesized genes were subcloned into the pCDNA3.4 vector for protein
expression in a CHO cell system. Subcloning into the pCDNA3.4 vector yielded the following
product: MGWSCIILFLVATATGVHS (signal peptide)-[Nanobody]-GGGGSHHHHHH, with a C-terminal
$6\times$His affinity tag.

\paragraph{VHH expression in CHO cells}\label{app:vhh-expression}
Transfection-grade plasmids were prepared in large quantities using a GenScript self-developed
plasmid extraction kit for CHO cell expression. Cells were maintained at $37\,^{\circ}\mathrm{C}$
with 5\% CO$_2$ on an orbital shaker. One day before transfection, the cells were seeded at an
appropriate density. On the day of transfection, DNA and Reagent (self-developed by GenScript) were
mixed at an optimal ratio and then added into cells ready for transfection. Approximately 24 hours
post transfection, Feed was added to each sample.

\paragraph{Protein purification and analysis}\label{app:vhh-purification}
Cell culture broth was centrifuged and followed by filtration. Filtered cell culture supernatant was
loaded onto an affinity purification column at an appropriate flowrate. After washing and elution
with appropriate buffers, the eluted fractions were pooled and buffer exchanged to the final
formulation buffer (PBS, pH 7.2). The purified protein was analyzed by SDS-PAGE and SEC-HPLC
analysis to determine the molecular weight and purity. The concentration was determined by A280
method.

\paragraph{Target protein preparation}\label{app:vhh-target-preparation}
EFNA5, PDGFR$\beta$, and S100A4 used the Fc-fusion constructs listed in
Table~\ref{tab:vhh_targets}. For BHRF1, commercially available proteins lacking alternative tags
were unavailable. Therefore, BHRF1 was expressed with a His-SUMO tag in HEK293 cells, followed by
tag cleavage.

\paragraph{Binding characterization of VHHs}\label{app:vhh-binding-characterization}
Initial screening of VHH designs for target protein binding was performed using BLI (Gator Bio).
For target proteins without His-tag (PDGFR$\beta$, EFNA5, BHRF1, and S100A4), His-tagged VHHs were
prepared at 5 $\mu$g/mL in PBSTA buffer. Target proteins were diluted to 1000 nM in PBSTA buffer.
Anti-His sensors (Gator Bio, Cat. No. 20-5066) were saturated with His-tagged VHHs. The BLI assay
sequence was as follows: 60 s baseline, 120 s ligand loading, 60 s post-loading baseline, 120 s
association phase, 300 s dissociation phase, and 30 s regeneration. Baseline-subtracted signals (in
nanometers) were calculated to prioritize VHHs for further characterization.

To determine apparent affinities of selected designs, 5 $\mu$g/mL His-tagged design VHHs prepared in PBSTA
were immobilized onto Anti-His sensor for 120 s. Serial dilutions of the target proteins (BHRF1:
from 2000 nM diluted to 31.25 nM, PDGFR$\beta$: from 3200 nM diluted to 50 nM, S100A4 and EFNA5 from 4000
nM diluted to 125 nM) were then allowed to associate with the immobilized ligand for 120 s,
followed by a dissociation step in PBSTA for 300 s. After background subtraction using buffer-only
(PBSTA) control curves, apparent $K_D$ values were determined using the 1:1 binding model in the Results \&
Analysis curve-fitting module.

\clearpage
\setcounter{table}{0}
\renewcommand{\thetable}{S\arabic{table}}
\section{Supplementary Tables}
\label{app:supplementary-tables}

Supplementary tables collect the target definitions, sampling settings, benchmark panels, and
experimental-validation reagents used in this study.

\begin{table}[H]
\centering
\scriptsize
\caption{Design tasks and corresponding hyperparameters.}
\label{tab:design_threeline}
\setlength{\tabcolsep}{3pt}
\begin{tabular}{p{2.2cm}p{6.5cm}p{3.2cm}p{2.0cm}}
\toprule
\textbf{Design task} & \textbf{Loss weight} & \textbf{Stage iterations} & \textbf{Learning rate} \\
\midrule
Minibinder &
  Intra-chain binder contact loss = 1 \newline
  Inter-chain target--binder contact loss = 1 \newline
  Binder PAE loss = 1 \newline
  Amino-acid composition loss = 5 &
  Stage 0 = 0 \newline
  Stage 1 = 0 \newline
  Stage 2 = 80 \newline
  Stage 3 = 5 &
  1 \\
\addlinespace
VHH &
  Paratope loss = 1 \newline
  ipTM loss = 10 \newline
  Binder--target interface PAE loss = 1 \newline
  CDR1-restricted IgLM loss = 1.5 \newline
  CDR2-restricted IgLM loss = 3.5 \newline
  CDR3-restricted IgLM loss = 5 &
  Stage 0 = 0 \newline
  Stage 1 = 0 \newline
  Stage 2 = 80 \newline
  Stage 3 = 5 &
  0.1 \\
\addlinespace
Cyclic peptide &
  Intra-chain binder contact loss = 1 \newline
  Inter-chain target--binder contact loss = 1 \newline
  Binder PAE loss = 1 \newline
  Amino-acid composition loss = 5 &
  Stage 0 = 30 \newline
  Stage 1 = 60 \newline
  Stage 2 = 45 \newline
  Stage 3 = 5 &
  0.1 \\
\addlinespace
Small-molecule binder &
  Intra-chain binder contact loss = 1 \newline
  Inter-chain target--binder contact loss = 1 \newline
  $\alpha$-Helix loss = 0.01 \newline
  Beta loss = 0.15 \newline
  Isotropy loss = 0.3 \newline
  Amino-acid composition loss = 5 &
  Stage 0 = 30 \newline
  Stage 1 = 75 \newline
  Stage 2 = 80 \newline
  Stage 3 = 5 &
  0.1 \\
\bottomrule
\multicolumn{4}{l}{\rule{0pt}{3ex}\footnotesize * All design tasks were optimized using the same SGD optimizer.} \\
\end{tabular}
\end{table}

\begin{table}[H]
\centering
\scriptsize
\caption{Mini-binder experimental-validation sampling settings for the designs reported in Figure~\ref{fig:2}a.}
\label{tab:minibinder_validation_settings}
\setlength{\tabcolsep}{3pt}
\begin{tabularx}{\textwidth}{llYYl}
\toprule
Design target & PDB ID & Target chain/residues & Hotspots & Length range \\
\midrule
IFNAR2 & 3se4* & C34--232 & C46,\allowbreak C48,\allowbreak C76,\allowbreak C80,\allowbreak C98,\allowbreak C100,\allowbreak C133,\allowbreak C138,\allowbreak C186,\allowbreak C189,\allowbreak C190 & 90--150 \\
IL-7R$\alpha$ & 7opb & B36--231 & B50,\allowbreak B76,\allowbreak B81,\allowbreak B97,\allowbreak B103,\allowbreak B155,\allowbreak B158,\allowbreak B162,\allowbreak B183,\allowbreak B187,\allowbreak B211,\allowbreak B212 & 90--150 \\
PDGFR$\beta$ & 3mjg & X124--312 & X136,\allowbreak X138,\allowbreak X186,\allowbreak X246,\allowbreak X259,\allowbreak X264 & 90--150 \\
PD-L1 & 8znl & B20--133 & B50,\allowbreak B64,\allowbreak B118,\allowbreak B126 & 60--130 \\
\bottomrule
\end{tabularx}
\begin{flushleft}
\scriptsize * For IFNAR2 (PDB: 3se4), missing residues at the protein interface were repaired using homology modeling via SWISS-MODEL.
\end{flushleft}
\end{table}

\begin{table}[H]
\centering
\scriptsize
\caption{VHH experimental-validation sampling settings for the designs reported in Figure~\ref{fig:3}a. The length range lists HCDR1, HCDR2, and HCDR3, respectively.}
\label{tab:vhh_validation_settings}
\setlength{\tabcolsep}{3pt}
\begin{tabularx}{\textwidth}{llYYl}
\toprule
Design target & PDB ID & Target chain/residues & Hotspots & CDR length range \\
\midrule
BHRF1 & 2wh6 & A2--158 & A65,\allowbreak A72,\allowbreak A81,\allowbreak A88,\allowbreak A107 & 8,8,9--21 \\
EFNA5 & 2x11 & B32--171 & B101,\allowbreak B130,\allowbreak B132 & 8,8,9--21 \\
PDGFR$\beta$ & 3mjg & X124--312 & X136,\allowbreak X138,\allowbreak X270 & 8,8,9--21 \\
S100A4 & 5lpu & C2--90 & C45,\allowbreak C58,\allowbreak C78 & 8,8,9--21 \\
\bottomrule
\end{tabularx}
\end{table}

\begin{table}[H]
\centering
\scriptsize
\caption{Mini-binder benchmark targets. Hotspots are listed separately for each binder-length setting.}
\label{tab:minibinder_benchmark_targets}
\setlength{\tabcolsep}{3pt}
\begin{tabularx}{\textwidth}{llYYYY}
\toprule
Target & PDB ID & Target chain/residues & 75-aa hotspots & 120-aa hotspots & 160/200-aa hotspots \\
\midrule
IL-7R$\alpha$ & 7opb & B36--231 & B78,\allowbreak B100,\allowbreak B159 & B79,\allowbreak B104,\allowbreak B155,\allowbreak B159,\allowbreak B184,\allowbreak B212 & B79,\allowbreak B104,\allowbreak B155,\allowbreak B159,\allowbreak B184,\allowbreak B187,\allowbreak B212 \\
InsulinR & 4zxb & E6--155 & E64,\allowbreak E88,\allowbreak E96 & E34,\allowbreak E37,\allowbreak E65,\allowbreak E86,\allowbreak E142,\allowbreak E147 & E9,\allowbreak E19,\allowbreak E34,\allowbreak E37,\allowbreak E65,\allowbreak E86,\allowbreak E142,\allowbreak E147 \\
PD-L1 & 8znl & B20--133 & B57,\allowbreak B116,\allowbreak B124 & B59,\allowbreak B70,\allowbreak B74,\allowbreak B120,\allowbreak B122,\allowbreak B126 & B30,\allowbreak B59,\allowbreak B70,\allowbreak B74,\allowbreak B107,\allowbreak B120,\allowbreak B122,\allowbreak B126 \\
TrkA & 1www & X282--382 & X294,\allowbreak X296,\allowbreak X333 & X287,\allowbreak X304,\allowbreak X306,\allowbreak X334,\allowbreak X339,\allowbreak X347 & X287,\allowbreak X291,\allowbreak X294,\allowbreak X296,\allowbreak X304,\allowbreak X306,\allowbreak X327,\allowbreak X329,\allowbreak X334,\allowbreak X339,\allowbreak X347 \\
CD3D & 1xiw & B2--72 & B6,\allowbreak B17,\allowbreak B38 & B4,\allowbreak B13,\allowbreak B17,\allowbreak B29,\allowbreak B32,\allowbreak B41 & B4,\allowbreak B9,\allowbreak B13,\allowbreak B17,\allowbreak B29,\allowbreak B32,\allowbreak B41 \\
EGFRc & 1mox & A301--500 & A384,\allowbreak A436,\allowbreak A468 & A384,\allowbreak A406,\allowbreak A411,\allowbreak A418,\allowbreak A463,\allowbreak A468 & A318,\allowbreak A357,\allowbreak A384,\allowbreak A406,\allowbreak A411,\allowbreak A418,\allowbreak A463,\allowbreak A468 \\
EGFRn & 1mox & A6--162 & A45,\allowbreak A47,\allowbreak A127 & A16,\allowbreak A20,\allowbreak A93,\allowbreak A101,\allowbreak A128,\allowbreak A153 & A16,\allowbreak A20,\allowbreak A30,\allowbreak A93,\allowbreak A101,\allowbreak A128,\allowbreak A148,\allowbreak A153 \\
FGFR2 & 1djs & A251--362 & A289,\allowbreak A347,\allowbreak A350 & A285,\allowbreak A289,\allowbreak A311,\allowbreak A317,\allowbreak A352,\allowbreak A354 & A285,\allowbreak A289,\allowbreak A311,\allowbreak A317,\allowbreak A352,\allowbreak A354,\allowbreak A356,\allowbreak A360 \\
IGF1R & 5u8r & A1--144 & A8,\allowbreak A59,\allowbreak A83 & A28,\allowbreak A58,\allowbreak A108,\allowbreak A114,\allowbreak A136,\allowbreak A138 & A10,\allowbreak A28,\allowbreak A58,\allowbreak A79,\allowbreak A108,\allowbreak A114,\allowbreak A136,\allowbreak A138 \\
PDGFR$\beta$ & 3mjg & X124--312 & X207,\allowbreak X242,\allowbreak X267 & X136,\allowbreak X138,\allowbreak X186,\allowbreak X246,\allowbreak X259,\allowbreak X264 & X136,\allowbreak X138,\allowbreak X186,\allowbreak X245,\allowbreak X246,\allowbreak X257,\allowbreak X259,\allowbreak X264 \\
Tie2 & 2gy7 & B23--445 & B152,\allowbreak B161,\allowbreak B192 & B66,\allowbreak B68,\allowbreak B95,\allowbreak B161,\allowbreak B167,\allowbreak B197 & B61,\allowbreak B66,\allowbreak B68,\allowbreak B71,\allowbreak B92,\allowbreak B95,\allowbreak B101,\allowbreak B161,\allowbreak B162,\allowbreak B167,\allowbreak B197 \\
VirB8 & 4o3v & A90--231 & A124,\allowbreak A140,\allowbreak A152 & A134,\allowbreak A138,\allowbreak A142,\allowbreak A186,\allowbreak A189,\allowbreak A193 & A134,\allowbreak A138,\allowbreak A142,\allowbreak A186,\allowbreak A189,\allowbreak A193,\allowbreak A196,\allowbreak A200,\allowbreak A217 \\
\bottomrule
\end{tabularx}
\end{table}

\begin{table}[H]
\centering
\scriptsize
\caption{VHH benchmark targets. All designs used CDR length settings of HCDR1 = 8 aa,
HCDR2 = 8 aa, and HCDR3 = 9--21 aa.}
\label{tab:nanobody_benchmark_targets}
\setlength{\tabcolsep}{3pt}
\begin{tabularx}{\textwidth}{llYY}
\toprule
Target & PDB ID & Target chain/residues & Hotspots \\
\midrule
IL13 & 1ijz & A1--113 & A11,\allowbreak A14,\allowbreak A15,\allowbreak A101,\allowbreak A107,\allowbreak A108 \\
HNMT & 1jqd & A5--292 & A175,\allowbreak A221,\allowbreak A269,\allowbreak A280 \\
1433E & 2br9 & A3--232 & A176,\allowbreak A199,\allowbreak A219,\allowbreak A230 \\
BHRF1 & 2wh6 & A2--158 & A65,\allowbreak A72,\allowbreak A81,\allowbreak A88,\allowbreak A107 \\
EFNA1 & 3hei & B18--149 & B46,\allowbreak B90,\allowbreak B114 \\
CCL2 & 4dn4 & M9--69 & M28,\allowbreak M39,\allowbreak M55,\allowbreak M61 \\
IL20 & 4doh & A24--176 & A62,\allowbreak A68,\allowbreak A160,\allowbreak A167,\allowbreak A171 \\
CSF1 & 4wrm & B2--146 & B10,\allowbreak B55,\allowbreak B56,\allowbreak B78,\allowbreak B86 \\
MZB1 & 7aah & A39--185 & A71,\allowbreak A86,\allowbreak A144,\allowbreak A147 \\
LIF & 7n0a & C10--181 & C14,\allowbreak C20,\allowbreak C127,\allowbreak C135 \\
XIAP & 1g3f & A263--343 & A307,\allowbreak A323,\allowbreak A324 \\
RSPO1 & 4bsp & A40--126 & A106,\allowbreak A110 \\
FCGR3B & 1fnl & A3--175 & A88,\allowbreak A90,\allowbreak A113 \\
UNG & 2hxm & A82--304 & A147,\allowbreak A158,\allowbreak A167 \\
DAF & 1h03 & P5--129 & P26,\allowbreak P47,\allowbreak P50 \\
\bottomrule
\end{tabularx}
\end{table}

\begin{table}[H]
\centering
\scriptsize
\caption{Cyclic-peptide benchmark targets.}
\label{tab:cyclic_peptide_benchmark_targets}
\setlength{\tabcolsep}{3pt}
\begin{tabularx}{\textwidth}{llYYl}
\toprule
Target & PDB ID & Target chain/residues & Hotspots & Peptide length \\
\midrule
MDM2 & 4hfz & A26--108 & A54,\allowbreak A58,\allowbreak A61 & 12--18 \\
MCL-1 & 2pqk & A172--197,\allowbreak A203--321 & A224,\allowbreak A227,\allowbreak A231,\allowbreak A235,\allowbreak A249,\allowbreak A253,\allowbreak A263 & 12--18 \\
GABARAP & 7zkr & A1--115 & A48,\allowbreak A50,\allowbreak A51,\allowbreak A52,\allowbreak A62,\allowbreak A65 & 12--18 \\
PD-L1 & 5o45 & A17--132 & A56,\allowbreak A115,\allowbreak A123 & 12--18 \\
KEAP1 & 2flu & A325--609 & A334,\allowbreak A380,\allowbreak A382,\allowbreak A415,\allowbreak A483,\allowbreak A530 & 12--18 \\
RagA:RagC & 6s6d & A4--303,\allowbreak B59--369 & B80,\allowbreak B226 & 12--18 \\
IL-23R & 5mzv & C24--C123 & C67,\allowbreak C100 & 12--18 \\
albumin & 1bj5 & A207--A385 & A212,\allowbreak A228,\allowbreak A325 & 12--18 \\
\bottomrule
\end{tabularx}
\end{table}

\begin{table}[H]
\centering
\scriptsize
\caption{Target proteins used for binder experimental validation.}
\label{tab:binder_targets}
\setlength{\tabcolsep}{3pt}
\begin{tabularx}{\textwidth}{llYllY}
\toprule
Protein & Company & Description & Cat. & Accession No. & Protein structure \\
\midrule
IFN-$\alpha$/$\beta$ R2 & ACRO & Human IFN-$\alpha$/$\beta$ R2 Protein, Fc Tag & IF2-H5255 & P48551-2 & IFNAR2(Ile27-Lys243)\_hFc \\
PDGFR$\beta$ & SinoBiological & Recombinant Human PDGF beta receptor/PDGFRB Protein (ECD), HPLC-verified & 10514-H02H1 & NP\_002600.1 & (Met1-Phe530)\_hFc \\
PD-L1 & SinoBiological & Recombinant Human PD-L1/B7-H1 Protein (ECD, hFc Tag), HPLC-verified & 10084-H02H & NP\_054862.1 & (Met1-Thr239)\_hFc \\
IL-7R$\alpha$ & ACRO & Human IL-7R$\alpha$/CD127 Protein, Fc Tag & ILA-H525a & P16871-1 & IL-7R$\alpha$(Glu21-Asp239)\_Fc(Pro100-Lys330) \\
\bottomrule
\end{tabularx}
\end{table}

\begin{table}[H]
\centering
\scriptsize
\caption{Target proteins used for VHH experimental validation.}
\label{tab:vhh_targets}
\setlength{\tabcolsep}{3pt}
\begin{tabularx}{\textwidth}{llYllY}
\toprule
Protein & Company & Description & Cat. & Accession No. & Protein structure \\
\midrule
BHRF1 & Oritop & BHRF1 & H2025111402 & P03182 & His-SUMO-(Met1-Ser165) \\
EFNA5 & SinoBiological & Recombinant Human Ephrin A5 Protein (hFc Tag) & 10192-H02H & NP\_001953.1 & (Met1-Asn203)\_hFc \\
PDGFR$\beta$ & SinoBiological & Recombinant Human PDGF beta receptor/PDGFRB Protein (ECD), HPLC-verified & 10514-H02H1 & NP\_002600.1 & (Met1-Phe530)\_hFc \\
S100A4 & SinoBiological & Recombinant Human S100A4 Protein (hFc Tag) & 10185-H01H & NP\_002952.1 & hFc\_(Met1-Lys101) \\
\bottomrule
\end{tabularx}
\end{table}

\begin{table}[H]
\centering
\scriptsize
\caption{TorchCraft runtime benchmark across sequence lengths on NPU (with/without fused attention, FA) and GPU (with/without cuequivariance). Batch size 2 is the recommended setting. Runtime is reported in minutes per design, obtained by dividing the total wall-clock time of a batch-size-2 run by 2. GPU timings are the mean of three independent runs.}
\label{tab:torchdesign_speed}
\setlength{\tabcolsep}{3pt}
\begin{tabularx}{\textwidth}{lllYYYYY}
\toprule
Code & Operator & Machine & 100 aa & 200 aa & 300 aa & 400 aa & 500 aa \\
\midrule
NPU & w/ FA & 910B (64G) & 3.1 & 3.75 & 6.23 & 11.29 & 16.56 \\
NPU & w/o FA & 910B (64G) & 3.0 & 4.13 & 8.63 & 14.64 & 33.47$^{\ast}$ \\
GPU & w/ cuequivariance & A100 (80G) & 3.92 & 4.27 & 4.54 & 5.85 & 7.59 \\
GPU & w/o cuequivariance & A100 (80G) & 4.76 & 7.96 & 13.08 & 21.65 & 33.3 \\
\bottomrule
\multicolumn{8}{l}{\footnotesize $^{\ast}$Batch size 1; batch size 2 exceeded memory (OOM).} \\
\end{tabularx}
\end{table}

\clearpage
\setcounter{figure}{0}
\renewcommand{\thefigure}{S\arabic{figure}}
\section{Supplementary Figures}
\label{app:supplementary-figures}

\begin{figure}[H]
  \centering
  \IfFileExists{figs-TD/Fig-SI/figS2-1.pdf}{
    \includegraphics[width=\textwidth,height=0.82\textheight,keepaspectratio]{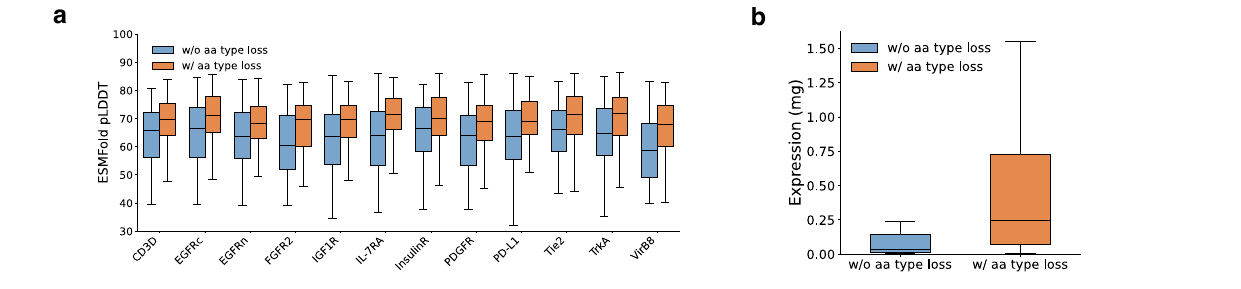}
  }{
    \fbox{\parbox[c][0.25\textheight][c]{0.92\textwidth}{
    \centering
    Missing figure file: figs-TD/Fig-SI/figS2-1.pdf.
    }}
  }
  \caption{
    \textbf{Sequence-composition regularization, predicted confidence, and
    observed minibinder expression.} (a) ESMFold pLDDT comparison for
    TorchCraft minibinders optimized with or without the composition
    regularizer across the 12-target benchmark. (b) PDGFR$\beta$ minibinder
    expression in two experimental rounds. The round incorporating composition
    regularization showed higher expression than the preceding unregularized
    round.
  }
  \label{fig:S1}
\end{figure}

\begin{figure}[H]
  \centering
  \IfFileExists{figs-TD/Fig-SI/figS2-3.pdf}{
    \includegraphics[width=\textwidth,height=0.82\textheight,keepaspectratio]{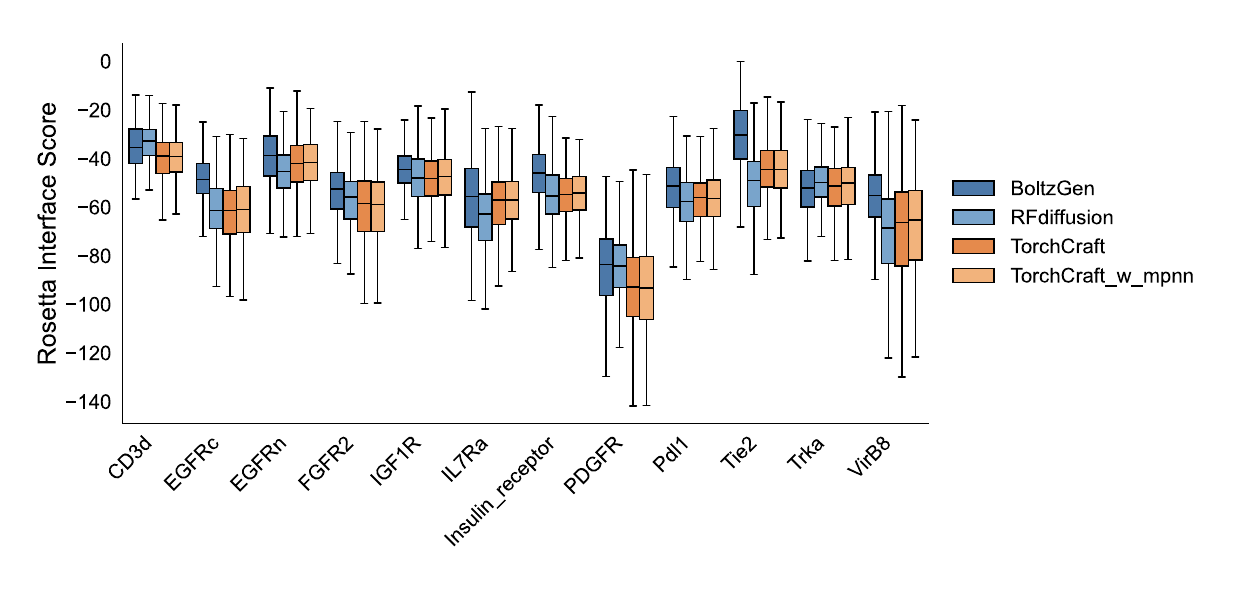}
  }{
    \fbox{\parbox[c][0.25\textheight][c]{0.92\textwidth}{
    \centering
    Missing figure file: figs-TD/Fig-SI/figS2-3.pdf.
    }}
  }
  \caption{
    \textbf{Rosetta interface-score benchmark for de novo minibinders.}
    Distribution of Rosetta interface scores for minibinder benchmark designs.
  }
  \label{fig:S2}
\end{figure}

\begin{figure}[H]
  \centering
  \IfFileExists{figs-TD/Fig-SI/figS3-3.pdf}{
    \includegraphics[width=\textwidth,height=0.82\textheight,keepaspectratio]{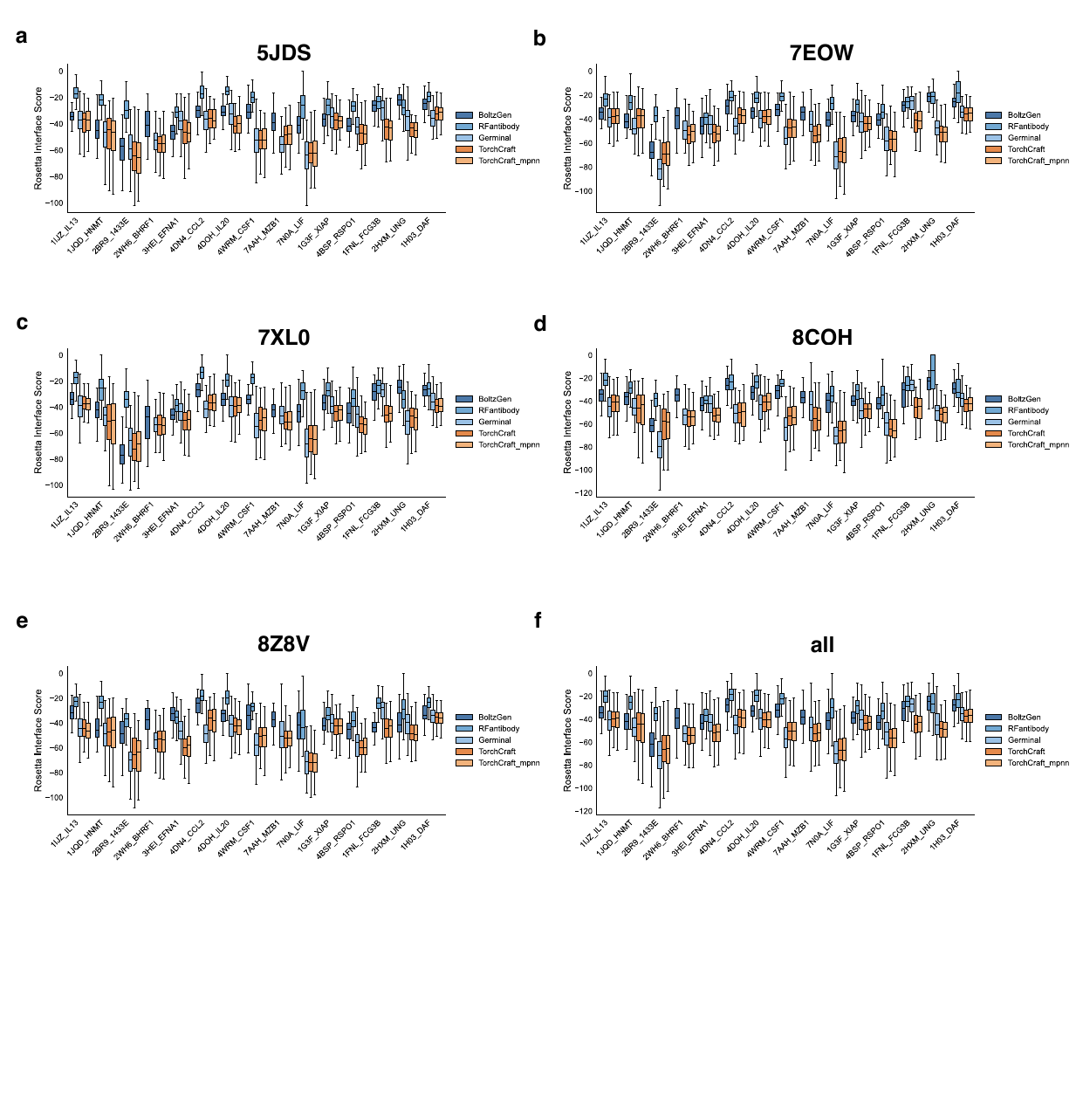}
  }{
    \fbox{\parbox[c][0.25\textheight][c]{0.92\textwidth}{
    \centering
    Missing figure file: figs-TD/Fig-SI/figS3-3.pdf.
    }}
  }
  \caption{
    \textbf{Rosetta interface-score benchmark for framework-conditioned
    nanobody designs.} Distribution of Rosetta interface scores for nanobody
    benchmark designs. Panels a--e show results for each of the five VHH
    frameworks, and panel f shows the average across frameworks.
  }
  \label{fig:S3}
\end{figure}

\begin{figure}[H]
  \centering
  \IfFileExists{figs-TD/Fig-SI/figS3-1.pdf}{
    \includegraphics[width=\textwidth,height=0.82\textheight,keepaspectratio]{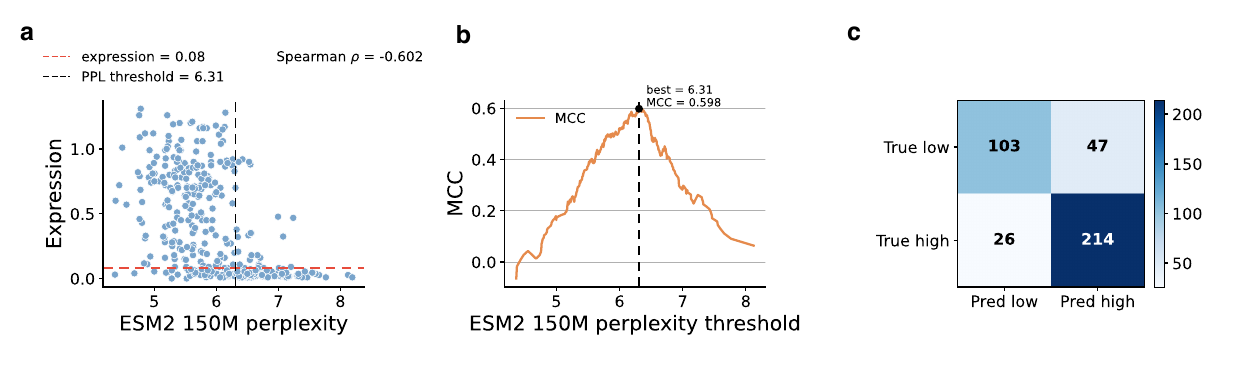}
  }{
    \fbox{\parbox[c][0.25\textheight][c]{0.92\textwidth}{
    \centering
    Missing figure file: figs-TD/Fig-SI/figS3-1.pdf.
    }}
  }
  \caption{
    \textbf{ESM2-150M perplexity retrospectively predicts VHH expression.}
    (a) Retrospective analysis of public expression measurements and
    accumulated TorchCraft expression data. ESM2-150M perplexity is correlated
    with expression, with lower perplexity generally associated with higher
    expression. The horizontal dashed line marks the high-expression cutoff of
    0.08 mg. (b) Threshold sweep for classifying high-expression designs by
    ESM2-150M perplexity. The selected threshold of 6.309 maximizes the Matthews
    correlation coefficient (MCC = 0.598; Pearson $r=-0.538$; Spearman
    $\rho=-0.602$), with designs below the threshold classified as
    high-expression candidates. (c) Confusion matrix obtained using the
    maximum-MCC perplexity threshold.
  }
  \label{fig:S4}
\end{figure}

\begin{figure}[H]
  \centering
  \IfFileExists{figs-TD/Fig-SI/figS3-2.pdf}{
    \includegraphics[width=\textwidth,height=0.82\textheight,keepaspectratio]{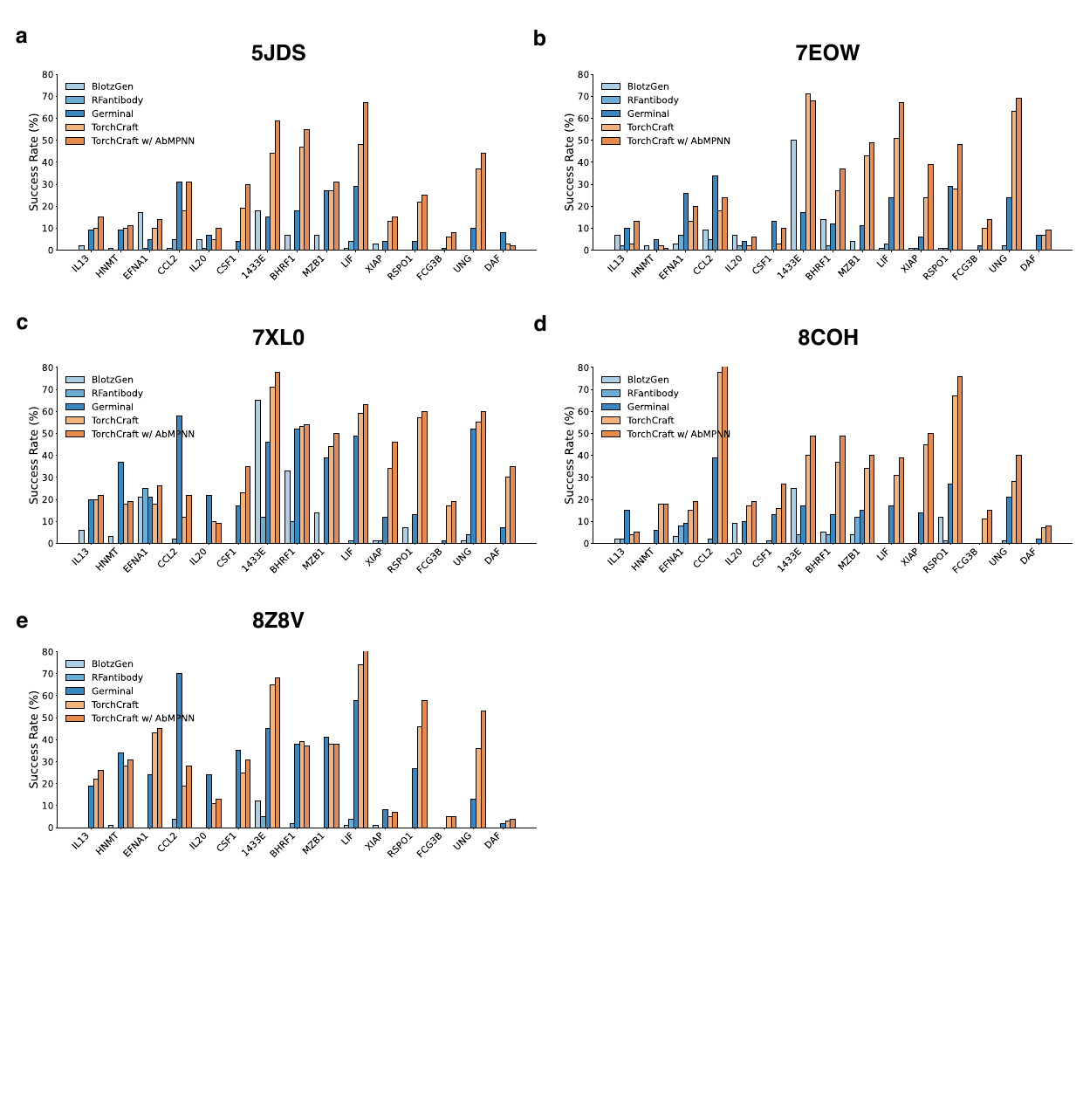}
  }{
    \fbox{\parbox[c][0.25\textheight][c]{0.92\textwidth}{
    \centering
    Missing figure file: figs-TD/Fig-SI/figS3-2.pdf.
    }}
  }
  \caption{
    \textbf{Framework-stratified VHH benchmark across five nanobody scaffolds.}
    TorchScore-based VHH design benchmark split by framework, showing how
    TorchCraft performance varies across five framework choices while using
    the same target panel and computational pass-rate criterion.
  }
  \label{fig:S5}
\end{figure}

\begin{figure}[H]
  \centering
  \IfFileExists{figs-TD/Fig-SI/figS4-1.pdf}{
    \includegraphics[width=\textwidth,height=0.82\textheight,keepaspectratio]{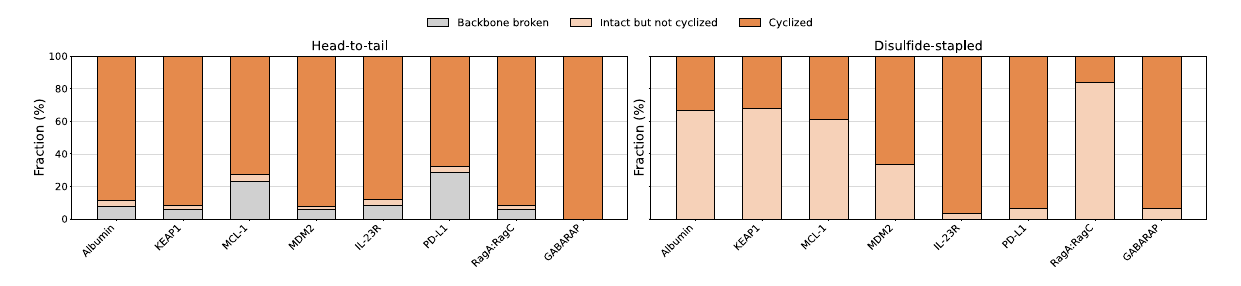}
  }{
    \fbox{\parbox[c][0.25\textheight][c]{0.92\textwidth}{
    \centering
    Missing figure file: figs-TD/Fig-SI/figS4-1.pdf.
    }}
  }
  \caption{
    \textbf{Cyclization outcomes for cyclic-peptide designs.} Fractions of
    successfully cyclized, intact-but-not-cyclized, and backbone-broken
    predictions across cyclic-peptide benchmark targets for the head-to-tail
    and disulfide-linked TorchCraft settings.
  }
  \label{fig:S6}
\end{figure}

\begin{figure}[H]
  \centering
  \IfFileExists{figs-TD/Fig-SI/figS4-2.pdf}{
    \includegraphics[width=\textwidth,height=0.82\textheight,keepaspectratio]{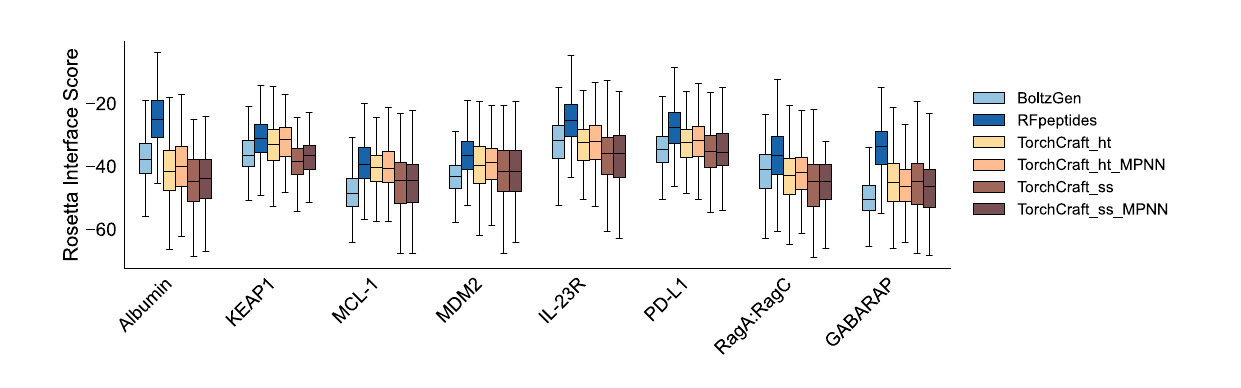}
  }{
    \fbox{\parbox[c][0.25\textheight][c]{0.92\textwidth}{
    \centering
    Missing figure file: figs-TD/Fig-SI/figS4-2.pdf.
    }}
  }
  \caption{
    \textbf{Rosetta interface-score benchmark for cyclic-peptide designs.}
    The cyclic-peptide benchmark from Figure~\ref{fig:4}c evaluated using
    Rosetta interface score as an orthogonal assessment of predicted
    target-binding quality.
  }
  \label{fig:S7}
\end{figure}

\begin{figure}[H]
  \centering
  \IfFileExists{figs-TD/Fig-SI/figS5-1.pdf}{
    \includegraphics[width=\textwidth,height=0.82\textheight,keepaspectratio]{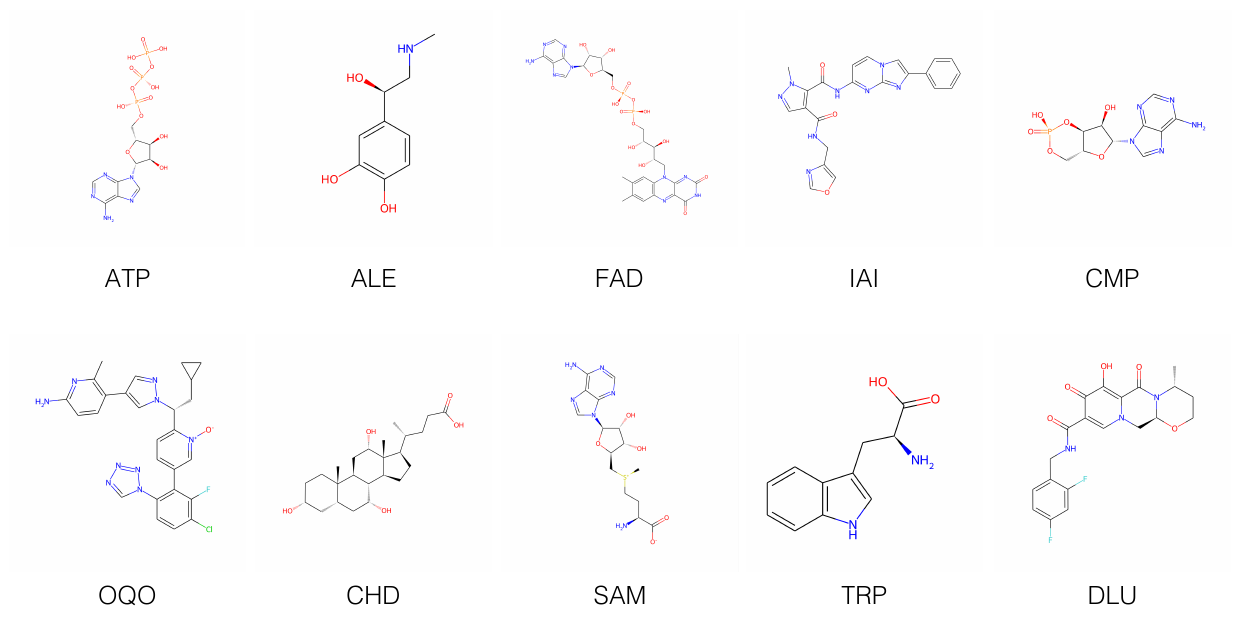}
  }{
    \fbox{\parbox[c][0.25\textheight][c]{0.92\textwidth}{
    \centering
    Missing figure file: figs-TD/Fig-SI/figS5-1.pdf.
    }}
  }
  \caption{
    \textbf{Small-molecule targets used in the SMI benchmark.} Benchmark panel
    of chemically diverse ligand targets used for small-molecule interaction
    design, including ALE, ATP, CHD, CMP, DLU, FAD, IAI, OQO, SAM, and TRP.
  }
  \label{fig:S8}
\end{figure}

\begin{figure}[H]
  \centering
  \IfFileExists{figs-TD/Fig-SI/figS5-2.pdf}{
    \includegraphics[width=\textwidth,height=0.82\textheight,keepaspectratio]{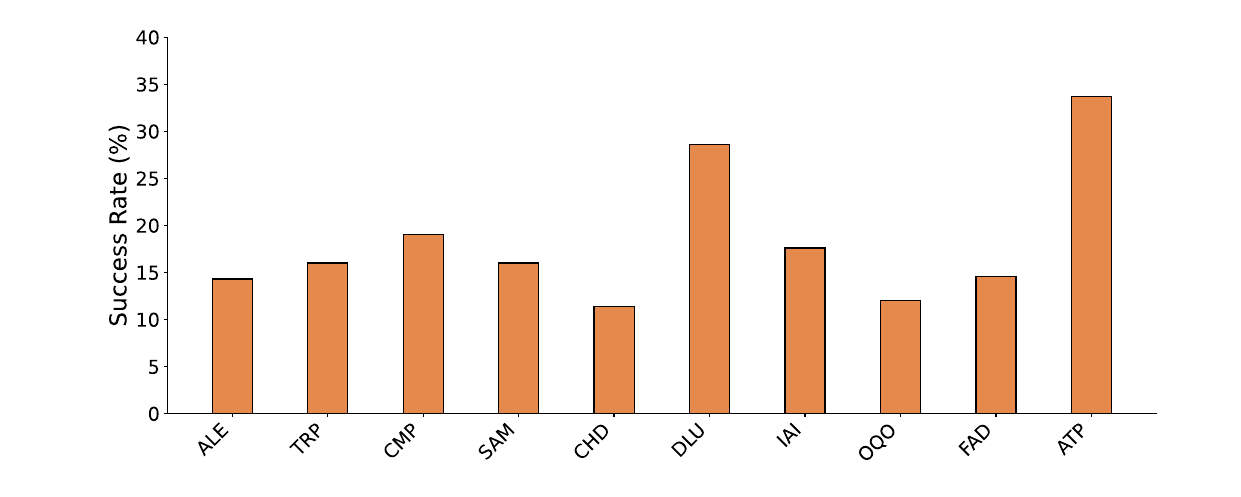}
  }{
    \fbox{\parbox[c][0.25\textheight][c]{0.92\textwidth}{
    \centering
    Missing figure file: figs-TD/Fig-SI/figS5-2.pdf.
    }}
  }
  \caption{
    \textbf{TorchScore computational pass rates in the small-molecule
    benchmark.} Pass-rate evaluation of TorchCraft ligand-conditioned protein
    designs across the 10-ligand panel using the TorchScore criterion.
  }
  \label{fig:S9}
\end{figure}

\end{document}